\def\gmode{} 

 \def\gdriver{dvipdfmx}

 \makeatletter 
 \@ifl@t@r{2014/06/01}\fmtversion{\def\gdriver{dviout}}{}

 \@ifl@t@r{2014/06/01}\fmtversion{}{}

\def\my@if@primitive#1{%
  \edef\my@tmpa{\string#1}\edef\my@tmpb{\meaning#1}%
  \ifx\my@tmpa\my@tmpb \expandafter\@firstoftwo
  \else \expandafter \@secondoftwo \fi}
\my@if@primitive\luatexversion{\def\gdriver{}}{}

\newcommand{\dtitle}[1]{\title{ \if \gmode \else
\color{red} Demo mode!\\
comment out \textbackslash def \textbackslash gmode\{demo\} at the header to include figures \color{black}\\
\fi
#1 }}
\ifx\pdfoutput\undefined
\else
 \def\gdriver{}
\fi

\documentclass[\gmode,\gdriver,10pt,twocolumn,letterpaper]{article}  

\usepackage[final]{cvpr}      
\usepackage{graphicx}
\usepackage{amsmath}
\usepackage{amssymb}
\usepackage{booktabs}

\usepackage{hyperref}

\usepackage[capitalize]{cleveref}
\crefname{section}{Sec.}{Secs.}
\Crefname{section}{Section}{Sections}
\Crefname{table}{Table}{Tables}
\crefname{table}{Tab.}{Tabs.}

\def\paperID{13} 
\def\confName{CVPR}
\def\confYear{2025}

\title{
Neural shape reconstruction from multiple views with static pattern projection
}

\author{%
Ryo Furukawa\textsuperscript{1}
\quad
Kota Nishihara\textsuperscript{2}
\quad
Hiroshi Kawasaki\textsuperscript{2}\\
\textsuperscript{1}Kindai University\\
\textsuperscript{2}Kyushu University\\
{\tt\small furukawa@hiro.kindai.ac.jp,
nishihara.kouta.869@s.kyushu-u.ac.jp,
kawasaki@ait.kyushu-u.ac.jp}
}

\usepackage{suffix,cprotect} 
\usepackage{siunitx} 
\usepackage{color}
\newcommand{\fnote}[1]{}
\newcommand{\fnoteII}[1]{{\color{blue} \bf #1 \color{black}}}

\newcommand{\knote}[1]{{\color{red} [K:] \bf #1 \color{black}}}
\newcommand{\knoteII}[1]{{\color{red} #1 \color{black}}}
\newcommand{\kcutcandidate}[1]{{\color{green} \bf #1 \color{black}}}
\newcommand{\bnote}[1]{{\color{red} #1 \color{black}}}
\newcommand{\mnote}[1]{{\color{blue} \bf #1 \color{black}}}
\newcommand{\review}[1]{{\color{blue} \bf #1 \color{black}}}
\newcommand{\onote}[1]{\color{blue} #1 \color{black}}

\newcommand{\kcut}[1]{}
\newcommand{\scut}[1]{}
\newcommand{\ocut}[1]{}
\newcommand{\jptext}[1]{{\color{magenta} \bf #1 \color{black}}}

\ifx\pdfoutput\undefined
\else
 \usepackage{CJKutf8}
 \renewcommand{\fnote}[1]{}
 \renewcommand{\knoteII}[1]{}
 \renewcommand{\fnoteII}[1]{}
 \renewcommand{\review}[1]{}
 \renewcommand{\kcutcandidate}[1]{}
 \renewcommand{\mnote}[1]{}
 \renewcommand{\bnote}[1]{}
 \renewcommand{\onote}[1]{}
 \renewcommand{\jptext}[1]{}
\fi

\ifx\argmin\undefined\newcommand{\argmin}{\mathop{\rm argmin}\limits}\fi
\ifx\argmax\undefined\newcommand{\argmax}{\mathop{\rm argmax}\limits}\fi
\ifx\bm\undefined\newcommand{\bm}[1]{\mbox{\boldmath{$#1$}}}\fi
\ifx\um\undefined\newcommand{\um}[1]{{\SI{#1}{\micro \metre}}}\fi 

\ifx\etal\undefined\newcommand{\etal}{{\it et al. }}\fi
\ifx\ie\undefined\newcommand{\ie}{{\it i.e.}}\fi
\ifx\eg\undefined\newcommand{\eg}{{\it e.g.}}\fi
\ifx\gt\undefined\newcommand{\gt}{$\textgreater$}\fi
\ifx\lt\undefined\newcommand{\lt}{$textless$}\fi
\ifx\aka\undefined\newcommand{\aka}{{\it a.k.a.}}\fi

\ifx\figref\undefined\newcommand{\figref}[1]{{Fig.\ref{#1}}}\fi
\ifx\Figref\undefined\newcommand{\Figref}[1]{{Figure~\ref{#1}}}\fi
\ifx\tabref\undefined\newcommand{\tabref}[1]{{Tab.\ref{#1}}}\fi
\ifx\Tabref\undefined\newcommand{\Tabref}[1]{{Table~\ref{#1}}}\fi
\ifx\equref\undefined\newcommand{\equref}[1]{Eq.(\ref{#1})}\fi
\ifx\Equref\undefined\newcommand{\Equref}[1]{Equation~(\ref{#1})}\fi
\ifx\secref\undefined\newcommand{\secref}[1]{Sec.\ref{#1}}\fi
\ifx\Secref\undefined\newcommand{\Secref}[1]{Section~\ref{#1}}\fi
\ifx\subsecref\undefined\newcommand{\subsecref}[1]{Sec.\ref{sec:#1}}\fi
\ifx\Subsecref\undefined\newcommand{\Subsecref}[1]{Section~\ref{sec:#1}}\fi

\graphicspath{{../}{./}{./figure-ba/}}

\begin{document}

\maketitle
\thispagestyle{empty}
\pagestyle{empty}

\begin{abstract}
   Active-stereo-based 3D shape measurement is crucial
   for various purposes, such as industrial inspection, reverse engineering,
   and medical systems, due to  its strong ability to accurately acquire
   the shape of textureless objects.
   Active stereo systems typically consist of a camera and a pattern projector,
   tightly fixed to
   each other, and 
    precise calibration between a camera and a projector is required, which in turn decreases the
   usability of the system. 
   If a camera and a projector can be freely moved during shape scanning process, it will
   drastically increase the convenience of the usability of the system.
   To realize it, we propose a technique to recover the shape of the target object
   by capturing multiple images while both the camera and the projector are in motion,
   and their relative poses are auto-calibrated
   by our neural signed-distance-field (NeuralSDF) using
   novel volumetric differential rendering technique.
   In the experiment, the proposed method is evaluated by performing 3D reconstruction
   using both synthetic and real images. 
\ifx
\fi
\end{abstract}


\begin{figure}[t]
        \centering
        \vspace{-0.3cm}
\begin{minipage}[b]{0.60\columnwidth}
    \centering
        \includegraphics[width=0.8\columnwidth]{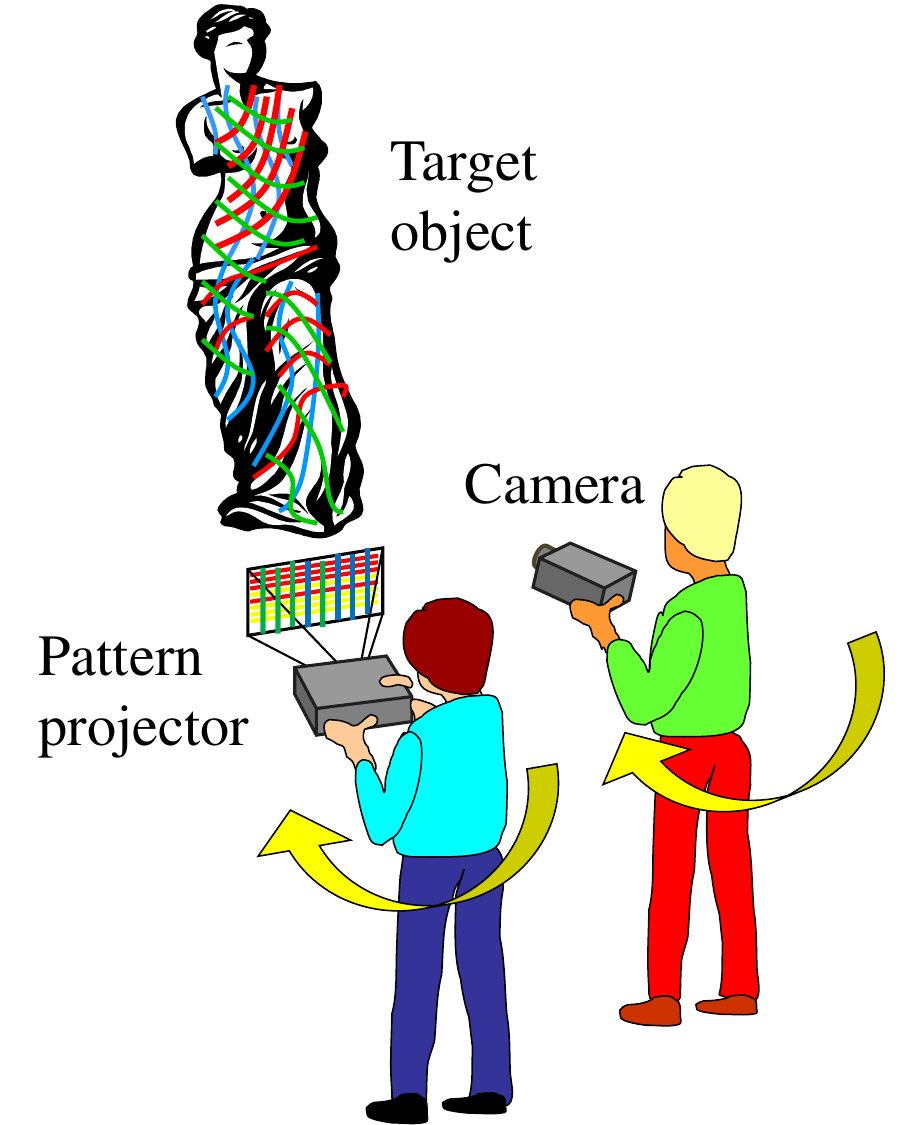}\\
        (a)
\end{minipage}
\begin{minipage}[b]{0.17\columnwidth}
    \centering
        \includegraphics[width=0.98\columnwidth]{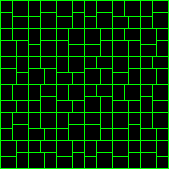}\\
        \includegraphics[width=0.98\columnwidth]{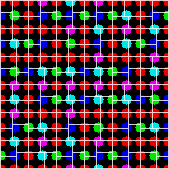}\\
        \includegraphics[width=0.98\columnwidth]{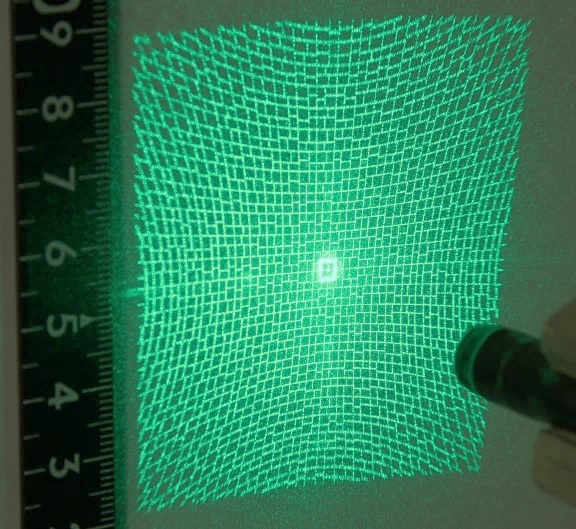} \\
        (b)
\end{minipage}
        \caption{
           (a)Active-stereo 3D scanning process with our system. A pattern projector and 
         a camera are freely moved during scan to reconstruct the entire shape of 
         the target object.
           (b:top) The projected pattern.
           (b:middle) The code information embedded into the pattern.
           (b:bottom) The actual pattern illuminated onto a plane.
        }
        \label{fig:system}        
        \vspace{-0.7cm}
     \end{figure}

     \jptext{
     ＝＝＝＝＝Reject＝＝＝＝＝＝
     ICPRのレビューの指摘事項
     DONE １．全体のプロセスが長い。（Fig.2とFig.3を足したような）全体を説明するOverviewの図があると分かりやすい。（その中で、もっと説明的な記述やラベルがあるとより分かりやすい）
     DONE(written in introduction) ２．画像は同時に撮影されるのか、個別に撮って後でICPするのかどうか、はっきりしない→後者
     DONE ３．Qualitativeな実験はあるが、Qualitativeな実験が無い→Experiment done 
     DONE ４．Ablation studyも無い（一部あるが、良く分からない）
     DONE ５．関連研究は、提案手法とのつながりが良く分からない。例えば、3Dレジストは何の関係がある？
     DONE by removing Sec.3.3 ６．U-NetやGCNで、CG画像を使って学習していると書いてあるが、How？良く分からない
     DONE ７．3.3節で、パターン画像からパターン画像をレンダリングする理由・どうやって、ということが不明
     Done ８．幾つかの変数が説明されてない。\mathcal{S}、　function f、t_i is Eq (6)、\mathbf{d}
     ９．Section 3.3の先頭は何書いてあるのか意味不明
     DONE １０．Last part on Section 3.2 refers to Fig 1(d) and 1(e), but Figure 1(e)
     DONE by removing Sec.3.3 １１．CGが何の略かが書いて無い
     DONE １２．3.4節の、NeRFのリファレンスが無い
     DONE１３．Bold face z’ in Eq. 2 should probably be non-bold face z’
     DONE１４．R_{w\rightarrow p} in Eq 2 is not explained
     DONE１５．In Eq (5) and (8) I would guess that the index of the first alpha should be j and not i
     DONE１６．In Eq (9) L is used to denote the st of camera poses. However, it also seems to be used to denote the loss L(f,L,M)
     DONE by replacing images １７．In Fig 5 (a,b) should probably be (a,d)
     ＝＝＝＝＝＝＝＝＝＝＝
     DONE １８．SLAMとの関連がない。SLAMペーパーだと過去研究が山程ある。
     DONE by removing Fig 2 and Sec.3.3 １９．Fig.2で、U-netが適用と書いてあるが、、、何のことか良く分からない
     DONE by removing Fig 2 and Sec.3.3 ２０．アウトライア処理がペーパーのStrengthと書いてあるが、実験結果で示して欲しい。特に[7]と比較して
     DONE ２１．実験で、weights w_{coord} とw_{pattern}の値がどんなだったか書いてない。
     ＝＝＝＝＝SR＝＝＝＝＝＝
     DONE ２２．比較やベンチマークとの違いが無い→Experiment done 
     ２３．SOTAとの比較がない→Experiment done 
     CUT ２４．Ablation studyが無い→４と同じ
     }
     
     \section{Introduction}
     \ifx
     A method using active light sources for stereo is a promising candidate for achieving 3D reconstruction and depth annotation,
     because of its strong ability to acquire
     the shape of textureless objects accurately.
     In particular, the one-shot scan techniques
     that exclusively relies on a single image
     with a static pattern,
     has gained significant attention due to its ability
     to capture moving objects, such as human bodies or internal organs.
     \ifx
        For example, Lin \etal proposed a method to obtain 3D deformable objects
        by projecting colored random dots from an active light source.
        Furukawa \etal proposed a method to generate a two-dimensional pattern using a laser with a single wavelength to obtain a shape.
        Such a method can obtain depth information from the correspondence between the camera image and the pattern by triangulation, even if the object shape is a single color.
     \fi
     For example,
     Lin~\etal~\cite{lin2015tissue}
     proposed a method where
     colored random dots are projected
     from an active light source to obtain deformable objects. 
     Furukawa~\etal~\cite{furukawa2018wide} used a two-dimensional pattern
     formed with a single-wavelength laser,
     by which shape inside a stomach of a pig was obtained. 
     These active stereo methods utilize projected patterns
     onto the target surfaces
     to obtain correspondences between the camera image and the pattern,
     even if the object has few textures.
     
     One of the important open problems of such a system is a calibration
     between a projector and a camera, because the pattern for one-shot scan is static and fixed to the projector, which makes it impossible to use several 
      patterns to apply well-known calibration techniques.
     If an extrinsic parameters between a camera and a projector can be auto-calibrated, 
     the usability of the system will be greatly improved.
     In addition, if auto-calibration is possible, 
     either devices can be freely moved during scan,
     which significantly enhances the convenience of scanning process, 
     allowing measurement of  entire shape of a target object as shown in \figref{fig:system}.
     
     \ifx 
     Typically, these systems require precise calibration
     between the projector and the camera poses
     before initiating the scanning processes to reconstruct accurate 3D shapes. %
     %
     While such a calibration may not be difficult for individuals
     familiar with structured-light (SL) systems,
     such expertise is uncommon within ordinary laboratories.
     Consequently, even though many laboratories have both cameras and video projectors,
     they often encounter difficulties in utilizing these tools for 3D shape scanning.
     %
     %
     %
     If a camera and a projector are freely moved during shape measurement
     without the need for calibration process, 
     it significantly enhances the convenience of using structured light, 
     \ie, video projectors, for 3D scanning.
     This allows users to achieve robust and accurate shape measurements
     without requiring in-depth knowledge of structured light techniques.
     %
     %
     One promising example of such a 3D scan is in endoscopic systems,
     where 
     it is challenging to securely fix a camera and a projector together.
     \fi  
     

     In this paper, we propose a technique to scan the target object
     from multiple images taken while a camera and a projector are both in motion.
     The relative poses between the camera and the projector are
     auto-calibrated.
     Note that no synchronization is required between a camera and a projector, since 
     they are auto-calibrated only from a captured image.
     The relative poses between the cameras (frames) are then simultaneously 
     optimized using 
     initial poses estimated by another method, \ie ICP in this paper.
     \ifx
        Similar to the method of Furukawa~\etal, this method uses a laser pattern projector forming grid-like patterns
        that features node-wise modulation for positional encoding.
        We utilize the modulation codes of the grid points to predict the correspondence between the camera-acquired image and the projected pattern to
        find global correspondences without special patterns for epipolar constraint.
        The pattern also facilitates the estimation of high-density correspondence by using pixel-wise phase estimation by deep neural network.
        By using the global correspondences, auto-calibration can be done for each frame while the projector is in motion, 
        which allows not only 3D scan for each frame, but also consistent shape integration of multiple frames.
     \fi
     
     For auto-calibration, we assume that correspondences between a captured image and a 
     projected pattern are densely obtained by using a deep neural network based method 
     using grid-like patterns~\cite{furukawa2018wide}.
     By utilizing the global correspondences, auto-calibration for each frame to estimate initial parameters were performed~\cite{furukawa2020fully}.
     By using the estimated parameters, 3D shape for each frame can be reconstructed, which can 
     be used to conduct ICP to obtain initial poses between frames.
     \ifx   
     In our method, similar to the method of Furukawa~\etal~\cite{furukawa2018wide}, a laser pattern projector which forms grid-like patterns to
     find global correspondences only from a single image was used.
     By utilizing these global correspondences, auto-calibration for each frame to estimate initial parameters were performed, while the projector and the camera are in motion.
     \fi   
     \ifx
        In addition, to calibrate the positional information of the projector and camera, we utilize multiple images taken while the projector is in motion.
        In this work, we assume that the target scene are static,
        which imposes constraints on the projector and camera positions.
     \fi
     Since the auto-calibrated parameters as well as poses estimated by ICP include inevitable errors,
     we employ optimization techniques using volumetric differential rendering 
     technique to integrate all the shapes consistently by refining all the parameters.
     In the process, we render a mapping from the camera pixels to the projector coordinates onto the surface derived from NeuralSDF
     and estimate the projector and the camera positions simultaneously
     by minimizing the differences between the rendered images and the observations.
     
     In the experiment, the proposed method is evaluated by performing 3D reconstruction
     using both synthetic and real images obtained by a pattern projector and a wide-field-of-view camera.
     %
     %
     The contributions of this paper are summarized as follows.
     (1)
              We propose a method to reconstruct a consistent 3D shape using observation
              from multiple views under structured-light (SL) projection. 
              For this purpose, a differential renderer specialized for 
              SL systems has been proposed.
        (2)
              We utilize neural shape representation inspired by 
              NeuS \cite{wang2021neus}, although color representation is entirely 
              modified to fit to structured-light systems. 
              Accordingly, new loss functions utilizing
              both projector-camera correspondences and 
              pattern appearances are proposed.  
             \ifx
        \item The method utilizes differential rendering to optimize the relative poses of the projector solely based on the observation of the projected pattern, without the need to observe markers or other objects.
              \fi
              \ifx
        \item To evaluate the proposed method, we conducted 3D reconstruction
              and compared the results with a ground-truth shape obtained from X-ray CT.
              We used images obtained from a surgical robot,
              where the projector and the camera were attached to different robotic arms,
              where the relative positions between the camera and the projector
              were dynamically changed during the operation.
              \fi
                   \ifx
        \item We propose novel undistortion algorithm for DOE projector, which has
              unique distortion and cannot be solved by existing undistortion algorithms.
              \fi
     (3)
        Thanks to joint optimization of calibration and consistent shape, shape
              integration from multiple images captured by freely moving the projector
              and camera during scan was achieved to recover an entire shape of object.
    \fi 

Active light-based stereo methods are promising for 3D reconstruction, especially of textureless objects. One-shot scanning, which uses a static pattern projected in a single image, enables capturing moving targets or deformable with small texture, such as human bodies or internal organs. For example, Lin~\etal~\cite{lin2015tissue} used colored random dots, and Furukawa~\etal~\cite{furukawa2018wide} used a laser-based grid pattern to scan inside a pig’s stomach.
A key challenge in such systems is calibration between a camera and a projector. Since the pattern is fixed, conventional multi-pattern based calibration methods cannot be used. If the extrinsic parameters could be auto-calibrated, it would allow flexible scanning with freely moving devices, as illustrated in \figref{fig:system}.

We propose a method for scanning with both camera and projector in motion, without synchronization. Relative poses between the camera and the projector are auto-calibrated from dense correspondences between captured images and projected patterns estimated by a deep 
 neural network~\cite{furukawa2018wide}. 
 These are then used to reconstruct 3D shape for each frame, which can be used to conduct ICP to obtain initial poses between frames. 
Since calibration parameters and ICP results are not consistent through  the frames, integrated shape inevitably has large errors.
In this paper, we propose a method to refine all the parameters through volumetric differential rendering based optimization. 
 We minimize discrepancies between rendered and observed images by jointly estimating projector and camera poses. Experiments on synthetic and real data validate our approach.


The contributions of this paper are: 
     (1)
              We propose a method to reconstruct a consistent 3D shape using observation
              from multiple views under structured-light (SL) projection using 
              differential renderer. 
        (2)
              We utilize neural shape representation inspired by 
              NeuS \cite{wang2021neus}, although color representation is entirely 
              modified to fit to SL systems, where 
              new loss functions utilizing
              both projector-camera correspondences and 
              pattern appearances are proposed.  
             \ifx
        \item The method utilizes differential rendering to optimize the relative poses of the projector solely based on the observation of the projected pattern, without the need to observe markers or other objects.
              \fi
              \ifx
        \item To evaluate the proposed method, we conducted 3D reconstruction
              and compared the results with a ground-truth shape obtained from X-ray CT.
              We used images obtained from a surgical robot,
              where the projector and the camera were attached to different robotic arms,
              where the relative positions between the camera and the projector
              were dynamically changed during the operation.
              \fi
                   \ifx
        \item We propose novel undistortion algorithm for DOE projector, which has
              unique distortion and cannot be solved by existing undistortion algorithms.
              \fi
     (3)
        Thanks to joint optimization of calibration and consistent shape, shape
              integration from multiple images captured by freely moving the projector
              and camera during scan was achieved to recover an entire shape of object.
     
              \ifx
        \item Since the proposed method does not require any calibration, shape
              integration from multiple images captured by freely moving the projector and camera during the operation are optimized
              to recover a consistent shape of a wide area.
              \fi
     \knoteII{貢献を変更する。これまでは、Autocalib（UPS）、ICP（手動初期位置合わせ）、Merge（高周波形状消失）を逐次的に処理。今回はこれをまとめて処理！
        その結果、計算時間の短縮、手動処理の大幅削減、高周波形状維持、というメリットがある。さらに、SLに微分レンダラーを使った初めての研究でもある。}

     \jptext{
        ●3D内視鏡の重要性について書く．
     
        例：主要サイズの推定
     
        例：deep learningのデータセット作成
     
        近年，内視鏡画像による人体内部の3次元計測システムが注目されている．
        ただし，内視鏡が対象とする内臓などの表面において，
        十分なテクスチャ情報が得られないことも多く，
        passive stereoでは、まだ安定した計測が十分に出来ない場合が多い．
     
        これに対して、Furukawa etal は，active stereoを利用した3D内視鏡を提案した．
        彼らは，鉗子口を通る超小型のファイバー投光器からパターン光を投影し，
        内視鏡カメラで撮影した，
        そして，撮影されたパターン光を用いて，
        画像の各画素からパターンの2D座標へのマップを推定した．
        彼らは，このマップを用いて，光切断法により単フレームにおいてdenseな形状復元に成功した~\cite{ryo-f:WACV}。
        彼らのシステムでは，投光器は内視鏡に固定されていないため
        カメラとパターン投光器の間の相対位置は自己キャリブレーションで
        毎回求められる．
        ただし自己キャリブレーションの基準は
        エピポーラ拘束のみであり，
        その結果はスケーリングの不一致や，
        誤差による形状歪みを含む．
     }
     
     
     \ifx
        Recent years have seen significant advances
        in image-based 3D shape measurement technology.
        Many previously difficult techniques,
        such as monocular posture and depth estimation
        of everyday objects using deep learning models,
        or SfM techniques with high accuracy,
        have become available for practical use.
        Among these technologies,
        active stereo systems continue to be
        important measurement techniques,
        especially in situations
        where there is little data available for training,
        target surfaces have less textures,
        or objective measurement is required.
        For example, in the medical fields,
        there is a demand to measure the surface tissue of internal organs,
        which generally have little data and little texture,
        and correct scales and certain measurement accuracies
        are needed.
        For the purpose,
        shape measurement techniques
        by active stereo method
        are drawing attentions~\cite{furukawa2020fully}.
     \fi
     \fnoteII{[F:]CVPR23: 上段落は維持}
     
     \fnoteII{[F:]CVPR23:
        未校正の能動ステレオ法の形状測定において，統合された形状データを
        得るには，多数のステップが必要である．
        つまり，測定，Autocalib, フレームごとの3次元復元，
        ICP(手動初期位置合わせ), Mergeである．
        これらの手法を順番に行うことは，
        ユーザにとって大きな負担である．
        また，さまざまな誤差の蓄積を招きやすい．
        特に，プロジェクタとカメラのtriangulationのベースラインが
        小さかったり，
        プロジェクタのepipoleがカメラ視界の近くにある場合には，
        フレームごとの３次元復元に大きな歪みが生じることがある．
     
        このような問題を解決するために，
        multi-view optimizationを使うことが考えられる．
        つまり，
        復数フレームでのプロジェクタ，カメラ間の対応情報から，
        シーンの形状パラメータと，
        プロジェクタ，カメラの位置パラメータを同時に
        推定することで，
        Auto-calibration,
        ICP, Mergeの手続きが，まとめて実行される．
     }
     \ifx
        \knote{BA is common solution for passive SfM, but cannot be used for SL}
        In passive stereo, bundle adjustment methods
        have made significant contributions
        to improving the accuracies of passive shape reconstruction.
        Bundle adjustment compensates for erroneous camera parameters
        by optimizing the observation errors of feature points,
        eliminating outliers,
        and providing positional and scale consistencies across frames.
        One of the reasons for the success of bundle adjustment
        is that the observation error of the passive stereo method
        can be modeled as simple and optimizable forms
        of ``reprojection errors''.
        On the other hand,
        there have been few attempts to apply
        multi-frame optimization to active stereo.
        KinectFusion~\cite{newcombe2011kinectfusion} is a multi-frame optimization for
        RGB-D cameras,
        but it does not model the projector and camera
        of active stereo methods.
        Furukawa~\etal~\cite{furukawa2019simultaneous} attempted to adapt
        a sparse bundle adjustment method to active stereo,
        but they did not directly model active-stereo process.
     \fi

     \ifx
        in which a pattern is projected from an ultra-small fiber projector
        that passes through the instrument channel of a common endoscope is proposed~\cite{furukawa2018wide,furukawa2005uncalibrated}.
     \fi
     \ifx
        By using the captured image with projected pattern,
        they estimated a map from the camera pixels to 2D coordinates of the projected pattern.
        Using the map, dense shapes can be obtained
        from each of the frames by using the light sectioning method~\cite{furukawa2018wide}
        for both textured and textureless environments.
     \fi
     
     \jptext{
        ●SfMやSLAMによるスケール不一致の解消
     
        上記のような歪を解消する手法として、画像系列を計測し，
        複数フレーム間での
        スケーリング不一致や，誤差の補正を行うことが提案されている．
        さらに、こうして連続して復元された形状情報をマージすることで，広いシーンを復元することもできる。
        このようなtaskは，パッシブステレオによるシステムでは，SfMやSLAMと呼ばれる。
        一般的には、３次元シーン中の特徴点と，
        その特徴点をカメラに投影した2次元座標を
        多数取得し，
        特徴点の３次元位置と
        各フレームのカメラ位置を，
        再投影誤差をlossとして最適化する
        bundle adjustment法で解かれる。
     
     
     
        プロジェクタは，しばしば
        カメラと同様にモデル化されることから，
        active stereoシステムでも，
        系列の計測カメラと
        プロジェクタの位置を，
        bundle adjustmentと同じような方式で
        最適化することが可能に思えるかもしれない．
        しかし残念ながら，
        その方法は自明ではない．
        その理由は以下の通りである．
        (1)active stereoシステムでは，
        テクスチャの無い状況を想定する．
        このため，
        active stereoシステムの時系列データでは，
        異なるフレームの間での対応点情報が存在しない．
        一つのフレーム内では，プロジェクタ，カメラ間の
        対応点マップが高い密度で得られるが，
        異なるフレームにおける計測情報の間の関連性は，
        重なって計測された形状情報が3次元的にfitするという
        ことのみである．
        (2)上記のことから，
        active stereoシステムの時系列データでは，
        一つの３次元点に対応する2次元点は，
        一つのフレームにおけるカメラ画像，および投影パターン上での点であり，
        その数は２でしかない．
        これは，
        passive stereoシステムでは，
        一つの３次元点が多数のフレームで観測され，
        複数フレームのカメラ位置の間の強い拘束条件になることに対して，
        対照的である．
     }

     \ifx
        \knote{Applying common SfM to active SL is theoretically difficult.}
     
        Since  a projector is geometrically equivalent to pinhole camera,
        one may consider if it is possible to optimize all the camera and projector parameters by bundle adjustment (BA) algorithm for passive stereo.
        However, 
        it is not true because of the following reasons.
        (1) In the active stereo system,
        textureless environment is assumed, and thus,
        explicit correspondences between different frames cannot be obtained,
        which is the key constraint for BA to achieve stable optimization.
        (2) From the above fact, 
        the frame number of 2D points corresponding to a 3D point is just two between a projector and a camera in the same frame. With such a small number, BA is not expected to work properly. 
     \fi
     
     \ifx
        It is worth to notice that
        multi-frame global optimizations are often used
        to estimate both stereo parameters and ego-motion parameters for camera only stereo systems.
        It is achieved by reprojection minimization algorithm
        called bundle adjustment (BA),
        where
        3D feature points and the camera positions are optimized
        by minimizing 
        the sum of reprojection errors,
        \ie, differences
        between the 2D observations
        and the projections of the 3D points onto the camera images.
        %
        %
     
        Although a projector is geometrically equivalent to pinhole camera,
        it is impossible to apply BA to the camera and projector systems because,
        in the active stereo system,
        textureless environment is assumed, and thus,
        explicit correspondences between different frames cannot be obtained,
        which is the key constraint for BA to achieve stable optimization.
     \fi
     
     \jptext{
        ●active stereoシステムにおけるBA方法
     
        このような、active stereoシステムにおける長い時系列データを通して不変なものは、シーンの形状である。
        複数フレーム間での位置合わせを形状のみによって行うには，
        ICPアルゴリズムも良く用いられる．
        しかし，
        ICPアルゴリズムは，
        今回提案する内視鏡用Active stereoのように、
        フレーム間でのスケールの不一致や，
        自己キャリブレーション誤差による形状歪みを考慮してない．
        複数フレームの情報を有効に活用するには，
        異なるフレームにおける形状の不一致を
        最小化することが必要である．
     
     
        本論文では，プロジェクタ・カメラ間が固定されないactive stereoシステムにおける
        多フレームの全体最適化のために，
        新しいloss関数を定義する．
        さらに，定義されたloss関数による
        全体最適化を，inverse renderingによって行う方法を提案する．
        このシステムは，以下の技術要素からなる．
        (1)投光器によるパターン投光による，
        フレーム内でのdenseな、カメラ座標とプロジェクタ座標間のマッピングの推定．
        (2)アクティブステレオにおけるパターン投影を直接モデル化した
        新しいloss関数の提案．
        (3)inverse renderingによって上記loss関数を最小化することによる
        多フレーム全体最適化．
     
        以下のセクションでは，これらの要素について説明する．
     
        最後に実験で本手法の有効性を示す。
     }
     
     
     
     
     
     
     \ifx
        \knote{following really necessary?}
     
        To the best of our knowledge, this work is the first study
        that discusses how multi-view active stereo is different
        from multi-view passive stereo, and models
        the active stereo system using projection mapping for global optimization
        purposes.
        In the following sections,
        these contributions are explained and finally, we demonstrate the effectiveness of our method by real experiments using endoscopic systems.
     \fi
     
     \fnoteII{[F:]CVPR23: ここに目的を挿入？
        未校正の能動ステレオ法の形状測定において，統合された形状データを
        得るには，多数のステップが必要である．
        つまり，測定，Autocalib, フレームごとの3次元復元，
        ICP(手動初期位置合わせ), Mergeである．
        これらの手法を順番に行うことは，
        ユーザにとって大きな負担である．
        また，さまざまな誤差の蓄積を招きやすい．
        特に，プロジェクタとカメラのtriangulationのベースラインが
        小さかったり，
        プロジェクタのepipoleがカメラ視界の近くにある場合には，
        フレームごとの３次元復元に大きな歪みが生じることがある．
     
        このような問題を解決するために，
        multi-view復元を使うことが考えられる．
        我々は，
        復数のカメラ-投光器対応のデータについて，
        それらのカメラパラメータ，プロジェクタパラメータの自己校正，
        フレーム同士の位置合わせ，
        形状統合処理
        を，まとめて行う手法を開発した．
        この手法では，
        multi-view情報を利用して，
        各フレームのカメラ及びプロジェクタのパラメータの推定，
        対象のシーンの形状情報の推定が行われる．


        KinectFusion~\cite{newcombe2011kinectfusion} などの手法が
        提案されているが，
        KinectFusionはdepth sensorの値が確定しいている必要があり，
        プロジェクタの位置パラメータがuncalibratedである場合には
        この手法は利用できない．
     }
     
     \section{Related Works}
     \ifx
     For estimating the projector and the camera parameters,
     a typical approach is projecting patterns onto a
        camera calibration object 
        such as a grid-printed or white planes~\cite{liao2008calibration,yamauchi2008calibration,wei2008flexible,audet2009user,drareni2009geometric}.
     These methods are for pre-calibration, where
     the projector and the camera are fixed and calibrated before measurement.
     To calibrate the parameters under dynamic scenes,
     Furukawa~\etal proposed a 
     methods using special markers embedded into the projected patterns~\cite{furukawa2018wide}, and then, extended it without using any markers~\cite{furukawa2020fully}.
     We also follow the similar approach to estimate the initial parameters without using any special patterns in our method.
     
     To integrate large areas of 3D scene from multiple scans, tons of researches 
     have been done, such as SLAM for depth sensors and V-SLAM or SfM for rgb cameras~\cite{dtam,slam++,mahmoud2018live,chen2018slam,leonard2018evaluation}.
     Recently, non-rigid SLAM have been proposed,
     such as Song~\etal~\cite{song2018mis},
     Lamarca~\etal~\cite{lamarca2020defslam},
     and Zhou~\etal~\cite{zhou2021emdq}.
     These methods use 2D features for rgb cameras or 3D features for a depth sensors, and 
     thus, textures on the objects are required for rgb cameras or ICP-based algorithms are 
     required for depth sensors. 
     Since we project patterns onto the objects, 2D feature based approach cannot be 
     applied, whereas 
     ICP-based algorithms is possible to 
     apply to obtain the initial parameters~\cite{besl1992method,sparseicp_sgp13,Yang_2013_ICCV,combes2010efficient,sinko20183d}.
     
     \ifx
     3D reconstruction based on passive camera, \eg, SLAM or SfM,
     for endoscopic images have been researched in medical image analysis,
     such as Mahmoud \etal~\cite{mahmoud2018live},
     Chen~\etal~\cite{chen2018slam},
     and Leonard \etal~\cite{leonard2018evaluation}.
     Recently, non-rigid SLAM have been proposed,
     such as Song~\etal~\cite{song2018mis},
     Lamarca~\etal~\cite{lamarca2020defslam},
     and Zhou~\etal~\cite{zhou2021emdq}.
     These methods need 3D feature points, thus
     needs textures.
     \fi
     
     \ifx
     One simple solution for textureless scene is to attach a marker to the projector and track its movements \cite{ExcelsiusGPS,StrykerNAVI3i};
     however, this approach has severe limitations, as the markers must be placed within the field of view, which is not possible for many cases.
     \fi

     \ifx
     For 3D registration, 
     ICP algorithm has been used~\cite{besl1992method,sparseicp_sgp13,Yang_2013_ICCV,combes2010efficient,sinko20183d}.
     In this paper, our target is
     not only registration of multiple 3D scenes,
     but
     simultaneously correcting inter-frame
     inconsistencies by taking the observation model
     into account.
     For such a purpose,
     Furukawa~\etal~\cite{Furukawa:embc2023} proposed a modification of bundle adjustment
     for passive and active stereo systems.
     Their method does not directly model dynamics of active stereo
     observations and has often problems in convergence if a projector and a camera
     are not set front parallel configuration. 
     In contract, our technique is robust and stable be proposing noise filtering
     algorithm for correspondence map as well as efficient undistortion algorithm for pattern projector using
     diffractive optical element (DOE).
     \fi
     
     
        To optimize 3D shapes from multiple images, differentiable renderer is a technique to render synthetic images using a scene and camera
        parameters where
        gradient-based optimization w.r.t. the parameters can be processed.
        Mesh-based differentiable renderers~\cite{henderson2018learning,palazzi2018end,kato2018neural,liu2018paparazzi}
        have been proposed to optimize various scene parameters
        such as geometry, illumination, textures, or materials and intensively
        researched, however, there remain several open problems, such as occlusion, mesh topology handling, etc.
          Recently, voxel-based differential renderers
          draw much attention because of solving such problems by using 
          neural-network-based representations such as NeRF~\cite{mildenhall2021nerf}
          or NeuS~\cite{wang2021neus}.
     
Several studies have employed NeRF or Neural SDF incorporate with
structured light (SL)~\cite{Rukun:3DV2024,Chunyu2022DRSL,NFSL,Ichimaru:3DV2024}. For instance, Rukun et al. and Li et al. combined
Neural SDF with Graycode for handling SL
employing a conventional decoding strategy for image-topattern
correspondence~\cite{Rukun:3DV2024,Chunyu2022DRSL}. Shandilya et al. integrated
pattern projection into a volume rendering pipeline, separately
modeling direct and indirect illumination, albeit requiring
reference images without pattern projection, which
may limit practical applicability~\cite{NFSL}. Ichimaru et al. proposed
a generalized Neural SDF approach accommodating
various patterns and system configurations, assuming Lambertian
surfaces exclusively~\cite{Ichimaru:3DV2024}. 
However, none of these techniques addressed the arbitrary positioning between the projector and the camera during each scan. Additionally, while they all considered photometric loss, they did not account for pattern decoding-based loss, both of which are strengths of our method.
\fi 

\ifx
[AA] Rukun Depth Reconstruction with Neural Signed Distance Fields in Structured Light Systems, 3DV, 2024
[11] ActiveNeuS: Neural Signed Distance Fields for Active Stereo, 3DV, 2024.
[13] Multi-View Neural Surface Reconstruction with Structured Light, BMVC, 2022
[28] Shandilya, A., et.al., “Neural fields for structured lighting,” 2023.
\fi     
     

Estimating the projector and the camera parameters is often done via pre-calibration using known
targets~\cite{liao2008calibration,yamauchi2008calibration,wei2008flexible,audet2009user,drareni2009geometric}. To support dynamic scenes, Furukawa~\etal~\cite{furukawa2018wide,furukawa2020fully} proposed marker-based and markerless calibration from captured images. Our method follows this idea but without requiring special patterns.

To integrate multiple views, SLAM and SfM approaches~\cite{dtam,slam++,mahmoud2018live,chen2018slam} have been proposed, including non-rigid variants~\cite{song2018mis,lamarca2020defslam,zhou2021emdq}. These rely on RGB textures or depth sensors, making them unsuitable for structured-light (SL) projection. Instead, we use ICP~\cite{besl1992method,sparseicp_sgp13,Yang_2013_ICCV} for initial alignment.

Unlike conventional registration, we refine inter-frame consistency using an observation model. Prior work~\cite{Furukawa:embc2023} applied bundle adjustment for active stereo, but struggled with non-parallel configurations. We improve robustness via correspondence filtering and undistortion tailored for DOE projectors.

Recent methods combine SL with differentiable rendering and neural representations~\cite{Rukun:3DV2024,Chunyu2022DRSL,NFSL,Ichimaru:3DV2024}, leveraging NeRF or NeuS~\cite{mildenhall2021nerf,wang2021neus}. However, they assume fixed camera-projector setups or Lambertian surfaces, and ignore pattern-decoding-based loss. Our method handles arbitrary poses and integrates both photometric and decoding losses for accurate optimization.

     \section{3D reconstruction with neural shape representation from structured-light projection}
     \subsection{Overview}
     \ifx
     In this paper, 
     we propose a 3D reconstruction method using pattern projection 
     observed by multiple camera positions. 
     The input images are multiple images captured 
     with structured-light projection
     while moving either a camera or
     a pattern projector. 
     The overall algorithm is composed of two steps.
     In the first step, 
     structured-light pattern in the captured images is analyzed using 
     deep-learning models, estimating dense projector-camera 
     correspondences as described in \secref{sec:systemconfig}.
     In the second step, 
     neural-represented shape, 
     camera poses, 
     and projector poses are simultaneously optimized using
     differential volume renderer.  
     The rendering model is similar to 
     NeuS~\cite{wang2021neus}; 
     however, we render structured-light features 
     such as projector patterns and coordinates 
     instead of scene radiance itself, details are described in \secref{sec:differential}.
     \fi 

We propose a 3D reconstruction method using structured-light pattern projection observed from multiple camera positions. The input consists of images captured while either the camera or the projector is in motion.

The algorithm has two steps. First, structured-light patterns in the images are analyzed using deep-learning models to estimate dense projector-camera correspondences (\secref{sec:systemconfig}). Second, the shape (neural-represented), camera poses, and projector poses are jointly optimized using a differential volume renderer.

Our renderer follows NeuS~\cite{wang2021neus}, but instead of rendering radiance, we render structured-light features such as projector patterns and coordinates (details in \secref{sec:differential}).

     \subsection{Image capturing and dense correspondences acquisition} \label{sec:systemconfig}

     \ifx
     We assume a shape capturing system
     that consists of a camera 
     and a pattern projector
     as illustrated in \figref{fig:system}.
     The system is based on a similar approach proposed
     by Furukawa~\etal~\cite{furukawa2020fully}.
     The structured light illumination is generated
     by a diffractive optical element (DOE)
     incorporated into the pattern projector.
     %
     We use a grid pattern
     consisting of vertical and horizontal edges
     with small gaps as shown in \figref{fig:system}(b:top).
     The gaps represent five code symbols to identify
     camera-to-projector mapping as shown in
     \figref{fig:system}(b:middle),
     where the red dots mean that the vertical and horizontal edges
     does not have gaps,
     the green dots and blue dots have gaps
     between horizontal edges with different gap directions
     (green means ``the left is higher'' and blue means ``the right is higher''),
     and the cyan and magenta dots have gaps
     between vertical edges similarly.
     The actual patterns projected onto the object surface are shown in \figref{fig:system}(b:bottom).
     
     
     %
     \begin{figure}[t]
        \begin{center}
           \includegraphics[width=0.6\linewidth]{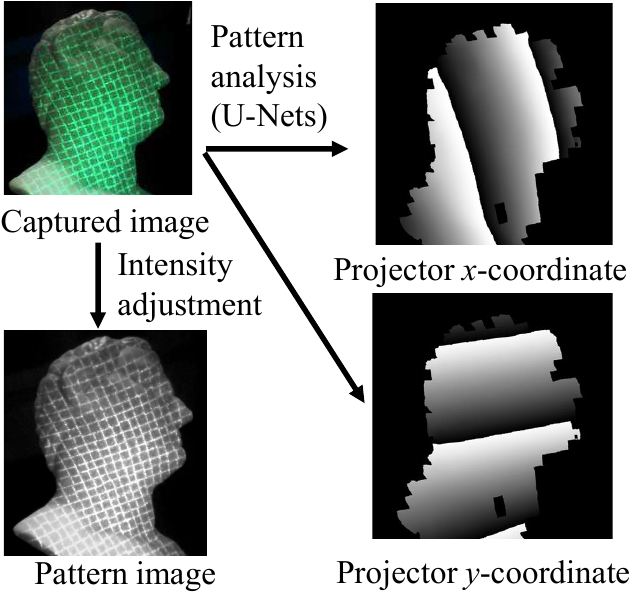}
           \caption{Overview of the preprocess. Projector coordinates and pattern 
            images are obtained from captured image using U-Nets (deep neural network).
     Projector coordinate images are visualized using modulus by 128 pixels.}
           \label{fig:reconstoverview}
        \end{center}
     \end{figure}


     By using the pattern projected images captured by the camera,  
     we obtain the camera-to-projector correspondences
     using the method proposed by Furukawa~\etal~\cite{furukawa2020fully}
     as shown in \figref{fig:reconstoverview}.
     In the input images 
     the grid pattern of 
     structured light can be seen. 
     We call these images 
     as `pattern images.'
     By using the correspondences of each frame, relative pose between the camera and the projector can be calibrated by using auto-calibration method~\cite{furukawa2020fully}.
     Then, by using the estimated parameters by auto-calibration, 3D shape for each frame can be reconstructed, which is
     used to conduct ICP to obtain initial poses between frames.
     Since the auto-calibrated parameters as well as poses estimated by ICP include inevitable errors,
     they are optimized by using volumetric differential rendering 
     technique in the next step.
     \fi 

We assume a shape capturing system consisting of a camera and a pattern projector, as shown in \figref{fig:system}, based on Furukawa~\etal~\cite{furukawa2020fully}. The projector uses a diffractive optical element (DOE) to generate structured-light patterns.
We employ a grid pattern with vertical and horizontal edges containing small gaps, as in \figref{fig:system}(b:top). These gaps encode five types of code symbols for identifying camera-to-projector mappings, illustrated in \figref{fig:system}(b:middle). The projected patterns appear as in \figref{fig:system}(b:bottom), and the captured images are referred to as “pattern images.”

Using Furukawa~\etal’s method~\cite{furukawa2020fully}, dense camera-to-projector correspondences are estimated from the pattern images (\figref{fig:reconstoverview}). These correspondences enable auto-calibration of the relative pose between the camera and projector for each frame.
Based on the estimated parameters, a 3D shape is reconstructed per frame. Initial inter-frame poses are computed via ICP. Since both the calibration and ICP contain inevitable errors, all parameters are refined in the next step using volumetric differential rendering.

     \begin{figure}[t]
        \begin{center}
           \includegraphics[width=0.6\linewidth]{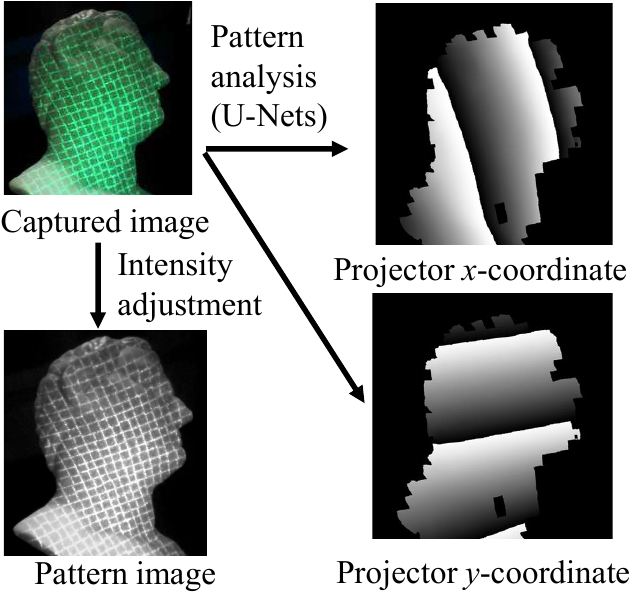}
           \caption{Overview of the preprocess. Projector coordinates and pattern 
            images are obtained from captured image using U-Nets (deep neural network).
     Projector coordinate images are visualized using modulus by 128 pixels.}
     \vspace{-5mm}
           \label{fig:reconstoverview}
        \end{center}
     \end{figure}

     \ifx   
     By using the pattern projector and the camera,  
     we obtain the camera-to-projector correspondences
     as shown in \figref{fig:reconstoverview}.
     In the input images 
     the grid pattern of 
     structured light can be seen. 
     We call these images 
     as `pattern images.'
     
     In the process, 
     U-Nets are applied to the pattern images captured by the camera to
     analyze the grid-like structure of the pattern.
     The U-Nets outputs two kinds of images;
     pixel-wise phase information,
     which represents the grid structure of the pattern,  
     and the code information.
     These U-Nets can be trained with CG and real images that simulates pattern projection.
     
     Then, the grid structure and code information  
     are converted to a grid graph (\figref{fig:reconstoverview}).
     This conversion can be done by segmentation similar to \cite{furukawa2020fully}.
     The graph is
     processed by a graph convolutional network (GCN),
     resulting in node-wise correspondences.
     The GCN can be trained with CG-generated graph data.
     
     Using the node-wise correspondences and
     the pixel-wise phase information,
     images representing camera-to-projector correspondence mapping are generated.
     For each pattern image, two images with x and y projector coordinates
     are obtained, where each pixel in camera-image space contains coordinates in pattern-image space (\figref{fig:reconstoverview}).
     We call these images as ($x$ or $y$) projector-coordinate images.
     
     The estimation of the correspondence map
     from pattern projection
     in one-shot method often introduce errors.
     In the method of \cite{furukawa2020fully},
     the codes in the projected pattern are analyzed,
     and global correspondences are estimated
     for each grid point of the grid pattern.
     If there is an error in the estimation of the global correspondence,
     the values of correspondence map
     in the grid region
     will be outliers.
     This causes negative impacts on shape optimization.
     
     To deal with this problem, 
     we implement outlier removal in the correspondence map using segmentation.
     As the correspondence estimation errors occur independently for each grid,
     the above outliers are
     isolated grid regions
     without continuity with the others.
     To utilize this,
     we segment the correspondence map
     into regions so that the discontinuous points of the x- or y-coordinates
     in the correspondence map
     become the boundaries of the region.
     Each region is either
     a correctly estimated correspondence
     of a continuous geometry,
     or an isolated outlier region.
     Outlier-isolated regions normally
     becomes small regions of about one or two grids,
     thus, regions with the number of pixels
     less than a threshold
     are removed and excluded from the correspondence map.
     Using the method,
     many of the errors in the correspondence map
     can be eliminated as shown in \figref{fig:reconstoverview}.

     
     


     \jptext{
        Bundle adjustmentでは，$P$の要素である一つの三次元点$p$が，
        複数のフレームで観測される場合が多い．
        単一点$p$が，複数のカメラで観測されることは，
        それらのカメラの間の位置関係の強力な拘束条件となり，
        Bundle adjustmentアルゴリズムの安定化に大きく寄与する．
     
        また，Bundle adjustmentアルゴリズムは，
        実際に存在する３次元点の集合$F$と，
        その観測である$O$の間の関係の
        「直接的なモデル化」であることに注意する必要がある．
        つまり，$F$の３次元点は，現実に存在し，
        $Projection(F,C)$は，
        $C$を動かしたときに予想される$F$の観測位置
        観測値$O$の予想$\tilde{O}$である．
     }



     \jptext{
        Active stereoにおける観測値は，
        カメラで観測された，カメラ画像上の２次元点$p_c$から，
        プロジェクタで投影されるパターン画像上での２次元点$p_p$への
        マッピングである(\figref{fig:observation-proj-cam})．
        この関係の，
        staticな幾何学配置は，
        passive stereoのBundle adjustmentと，似ているように
        見えるかもしれない．
        しかし，
        物理的な現象は異なる．
        active stereoでは，
        パターンの$q_1$から投影された光線が，
        シーンの３次元形状$S$に衝突し，
        その位置$s_1$がカメラの$p_c$で観測されることを意味する．
        この時，
        プロジェクタのパラメータを動かした時，
        $p_p$の光線が$S$に衝突した点$s_2$は，
        $s_1$とは異なる．
        カメラの内部及び外部パラメータの集合を
        $C=\{c_1,\cdots,c_n\}$とし，
        プロジェクタの内部及び外部パラメータの集合を
        $P=\{p_1,\cdots,p_m\}$とすると，
        Active stereoにおける観測値は
        $P$からのレイと$S$の交点が，$C$で観測されたものである．
     }

     \fi   
     
     \subsection{Differential volumetric rendering for structured-light projection}\label{sec:differential}

     \ifx
     We reconstruct a consistent 3D scene from multiple images captured with static structured-light pattern onto the scene. 
     To achieve the goal, 
     neural representation of signed-distance-field (NeuralSDF) is used as same as NeuS~\cite{wang2021neus}. 
     To utilize information from active pattern projection,
     we render two kinds of images, which are pattern images
     and projector-coordinate images.
     These images can be rendered using NeuralSDF, 
     camera and projector poses, and the pattern image that is projected.
     The rendering is done with differential volume renderer, with the same way as NeRF~\cite{mildenhall2021nerf} or NeuS~\cite{wang2021neus}.
     
     
     %
     %
     %
     
     First, we describe the method for rendering projector-coordinate images.
     A combination of an $x$-projector-coordinate image and a $y$-projector-coordinate image 
     represent a 2D-to-2D mapping
     \begin{equation}
        H\,:\, \mathbb{R}^2 \mapsto \mathbb{R}^2 \,;\, (r_x, r_y) \mapsto (q_x, q_y)
     \end{equation}
     from camera pixels $(r_x, r_y)$
     to projector pixels $(q_x, q_y)$.
     A combination of $x$ and $y$ projector-coordinate images 
     represents a mapping $H$.


     
     We use a differential volume renderer similar to NeuS
     to render projector-coordinate images, \ie, $H(r_x, r_y)$.
     In original NeuS, 
     a neural 3D field is used to represent a signed distance field, 
     and also for color intensities of a light field. 
     However, in this paper, 
     we use a neural 3D field only for a signed distance field, 
     and not for colors.
     Instead, 
     we use $\bf{c}$, which maps
     a 3D point $\bf{p}$ to
     2D projector coordinates as shown in \figref{fig:coordinaterendering}.
     The function 
     $\bf{c}$ is often used in 
     CG rendering to achieve `projection mapping.'
     
     $\bf{c}:\mathbb{R}^3  \rightarrow \mathbb{R}^2$
     is defined as

     \begin{align}
        \mathbf{c}(\mathbf{p}) &= \frac{1}{-z'}
        \begin{bmatrix}
           \alpha_x {x'} \\
           \alpha_y {x'} \\
        \end{bmatrix}
        +
        \begin{bmatrix}
           \beta_x \\
           \beta_y \\
        \end{bmatrix},\\
        \text{where}
        \begin{bmatrix}
           x' \\
           y' \\
           z' \\
        \end{bmatrix}
        &= \mathbf{R}_{wp}~ \mathbf{p} +\mathbf{t}_{wp}.
        \label{eqn:projection}
     \end{align}
     $\mathbf{R}_{wp}$ and $\mathbf{t}_{wp}$ 
     are rotation matrix and translation vector
     from the world coordinates 
     to the projector coordinates. 
     
     We render $\bf{c}$ for an SDF-represented surfaces as shown 
     in \figref{fig:coordinaterendering}.
     
     The surface $\mathcal{S}$ is represented as a zero level set of SDF
     in the same way as NeuS.
     \begin{align}
        \mathcal{S}=\{\mathbf{x}\in \mathbb{R}^3 | f(\mathbf{x})=0\} \label{eqn:sdf}
     \end{align}

     \begin{figure}[t]
        \centering
        \includegraphics[width=0.9\columnwidth]{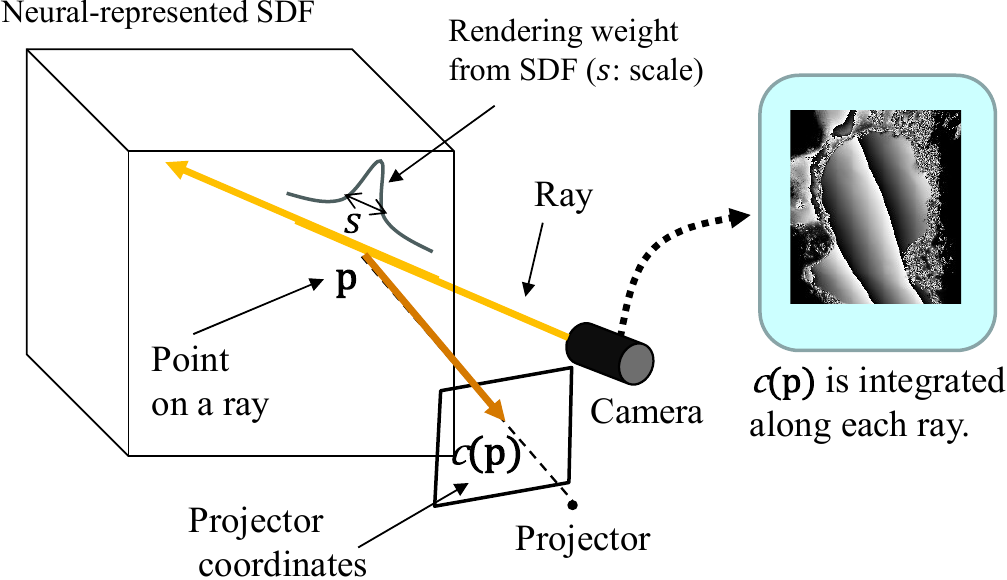}\\
        \caption{Rendering projector-coordinate images}
        \label{fig:coordinaterendering}
        \vspace{-0.3cm}
     \end{figure}
     
     We render `projector coordinates' by using the following rendering
     method.
     A ray from a camera can be represented by camera optical center
     $\mathbf{o}$ and ray direction $\mathbf{v}$, as
     $\{\mathbf{p}(t) \equiv \mathbf{o} + t\mathbf{v}| t \ge 0\}$.
     The rendered  value is as follows:
     \begin{align}
        \mathbf{C}(\mathbf{o}, \mathbf{v}) = \int_0^{+\infty} {w}(t)~\mathbf{c}(\mathbf{p}(t))~dt, \label{eqn:render}
     \end{align}
     where ${w}(t)$ is a weight function 
     that can be calculated from SDF $f$ in the same way as NeuS. 
     We use bold $\mathbf{C}$ because it is a 2D vector of $x$ and $y$ coordinates. 
     Since this equation renders 
     projector coordinates $\mathbf{c}(\mathbf{p}(t))$
     instead of radiance-field values, 
     we can render projector coordinates
     as shown in \figref{fig:coordinaterendering}.
     
     By discretizing Equation \eqref{eqn:render}
     with the same way as \cite{wang2021neus}, we use
     \begin{align}
        \hat{\mathbf{C}} = \sum_{i=1}^{n} ~\{\Pi _{j=1}^{i-1}(1-\alpha_j) \}\alpha_i ~ \mathbf{c}_i,
     \end{align}\label{eqn:renderdiscrete}
     where
     \begin{align}
        \alpha_i = \max (
        \frac{\Phi(s,f({\mathbf p}(t_i))) - \Phi(s, f({\mathbf p}(t_{i+1})))}
        {\Phi(s, f(\mathbf{p}(t_i)))},
        0). \label{eqn:renderdiscretealpha}
     \end{align}
     $alpha_i$ represents opacity of sampled voxels,
     $f$ is an SDF value from equation \eqref{eqn:sdf},
     $\Phi(s,x)=(1+e^{-sx})^{-1}$ is Sigmoid function with scale parameter $s$,
     and $\mathbf{c}_i$ is a sampled value
     of $\mathbf{c}(\mathbf{p}(t))$,
     $s$ represents `thickness' of the surface that controls the range of integration 
     around the zero-crossing surfaces. 
     Controlling of $s$ is discussed later.

     For rendering `pattern images',
     we use ${T}(\mathbf{c}(\mathbf{p}(t))$ instead of $\mathbf{c}(\mathbf{p}(t))$,
     where ${T}$ is a bilinear texture access of the projected pattern. 
     By replacing $\mathbf{c}(\mathbf{p}(t))$ with ${T}(\mathbf{c}(\mathbf{p}(t))$, the above rendering pipeline
     renders `pattern images.'
     Thus, pattern images can be rendered by
     \begin{align}
        P(\mathbf{o}, \mathbf{v}) = \int_0^{+\infty} {w}(t)~{T}(\mathbf{c}(\mathbf{p}(t)))~dt. \label{eqn:renderpattern}
     \end{align}
     We use non-bold $T$ and $P$ because it is a 1D intensity of texture image. 
     The discretized form is 
     \begin{align}
        \hat{P} = \sum_{i=1}^{n} \{\Pi _{j=1}^{i-1}(1-\alpha_j) \}\alpha_i~ {T}(\mathbf{c}_i).
     \end{align}
     \fi 

We reconstruct a consistent 3D scene from multiple images captured under static structured-light projection. To this end, we use NeuralSDF, following NeuS~\cite{wang2021neus}, but adapted for structured-light input.

To incorporate active projection information, we render two types of images: pattern images and projector-coordinate images. These are generated using NeuralSDF, along with the camera/projector poses and the projected pattern image, using a differential volume renderer similar to NeRF~\cite{mildenhall2021nerf} and NeuS~\cite{wang2021neus}.

We first describe the rendering of projector-coordinate images. A pair of $x$- and $y$-projector-coordinate images defines a 2D-to-2D mapping
     \begin{equation}
        H\,:\, \mathbb{R}^2 \mapsto \mathbb{R}^2 \,;\, (r_x, r_y) \mapsto (q_x, q_y)
     \end{equation}
     from camera pixels $(r_x, r_y)$
     to projector pixels $(q_x, q_y)$.
     A combination of $x$ and $y$ projector-coordinate images 
     represents a mapping $H$.

Unlike NeuS, which uses neural fields for both SDF and color, we only use it for the SDF. 
     Instead, 
     we use $\bf{c}$, which maps
     a 3D point $\bf{p}$ to
     2D projector coordinates as shown in \figref{fig:coordinaterendering}.
     The function 
     $\bf{c}$ is often used in 
     CG rendering to achieve `projection mapping.'
     \begin{align}
        \mathbf{c}(\mathbf{p}) &= \frac{1}{-z'}
        \begin{bmatrix}
           \alpha_x {x'} \\
           \alpha_y {x'} \\
        \end{bmatrix}
        +
        \begin{bmatrix}
           \beta_x \\
           \beta_y \\
        \end{bmatrix},\\
        \text{where}
        \begin{bmatrix}
           x' \\
           y' \\
           z' \\
        \end{bmatrix}
        &= \mathbf{R}_{wp}~ \mathbf{p} +\mathbf{t}_{wp}.
        \label{eqn:projection}
     \end{align}
where $\begin{bmatrix} x' \ y' \ z' \end{bmatrix} = \mathbf{R}_{wp} \mathbf{p} + \mathbf{t}_{wp}$ is the transformation from world to projector coordinates.

We render $\mathbf{c}$ over the surface represented by the SDF, as illustrated in \figref{fig:coordinaterendering}.

     \begin{figure}[t]
        \centering
        \includegraphics[width=0.8\columnwidth]{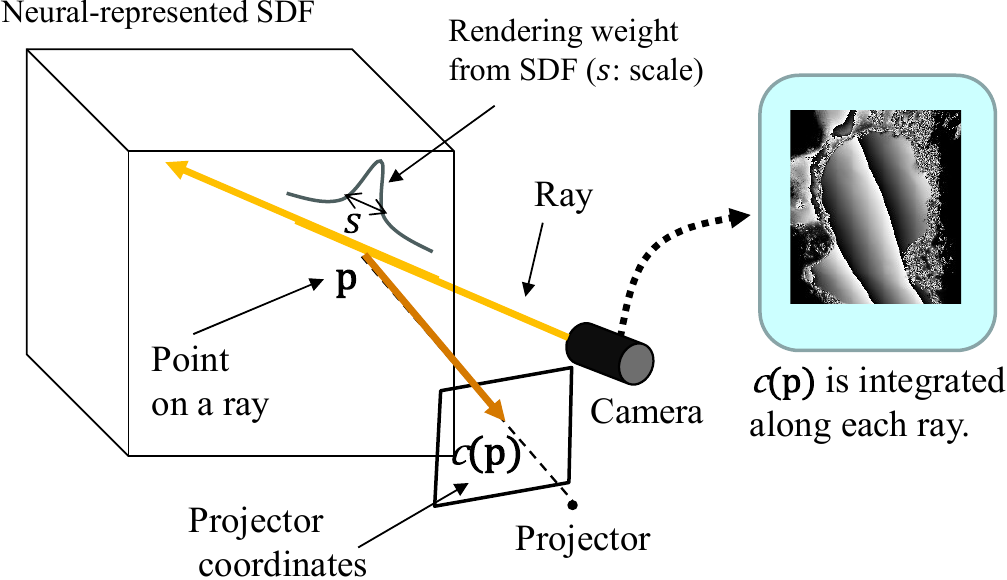}\\
        \caption{Rendering projector-coordinate images}
        \label{fig:coordinaterendering}
        \vspace{-0.3cm}
     \end{figure}

The surface $\mathcal{S}$ is defined as the zero level set of the signed distance field (SDF), following NeuS:
     \begin{align}
        \mathcal{S}=\{\mathbf{x}\in \mathbb{R}^3 | f(\mathbf{x})=0\} \label{eqn:sdf}
     \end{align}
A ray from the camera is parameterized as $\mathbf{p}(t) = \mathbf{o} + t\mathbf{v}$, where $\mathbf{o}$ is the optical center and $\mathbf{v}$ is the direction. 
Projector-coordinate images are rendered as:




     \begin{align}
        \hat{\mathbf{C}} = \sum_{i=1}^{n} ~\{\Pi _{j=1}^{i-1}(1-\alpha_j) \}\alpha_i ~ \mathbf{c}_i,
     \end{align}\label{eqn:renderdiscrete}

with

     \begin{align}
        \alpha_i = \max (
        \frac{\Phi(s,f({\mathbf p}(t_i))) - \Phi(s, f({\mathbf p}(t_{i+1})))}
        {\Phi(s, f(\mathbf{p}(t_i)))},
        0). \label{eqn:renderdiscretealpha}
     \end{align}

where $\mathbf{C}$ is a 2D vector, $w(t)$ is a weight function derived from the SDF and 
$\Phi(s, x) = (1 + e^{-s x})^{-1}$ is a sigmoid with scale parameter $s$, controlling surface thickness.

To render pattern images, we replace $\mathbf{c}$ with bilinear texture lookup $T$, resulting in:



     \begin{align}
        \hat{P} = \sum_{i=1}^{n} \{\Pi _{j=1}^{i-1}(1-\alpha_j) \}\alpha_i~ {T}(\mathbf{c}_i).
     \end{align}

     \subsection{Optimization strategy}\label{sec:optimization}
     \ifx
\begin{figure*}[t]     
     \begin{minipage}[b]{0.8\hsize}
        \centering
        \includegraphics[width=1.0\linewidth]{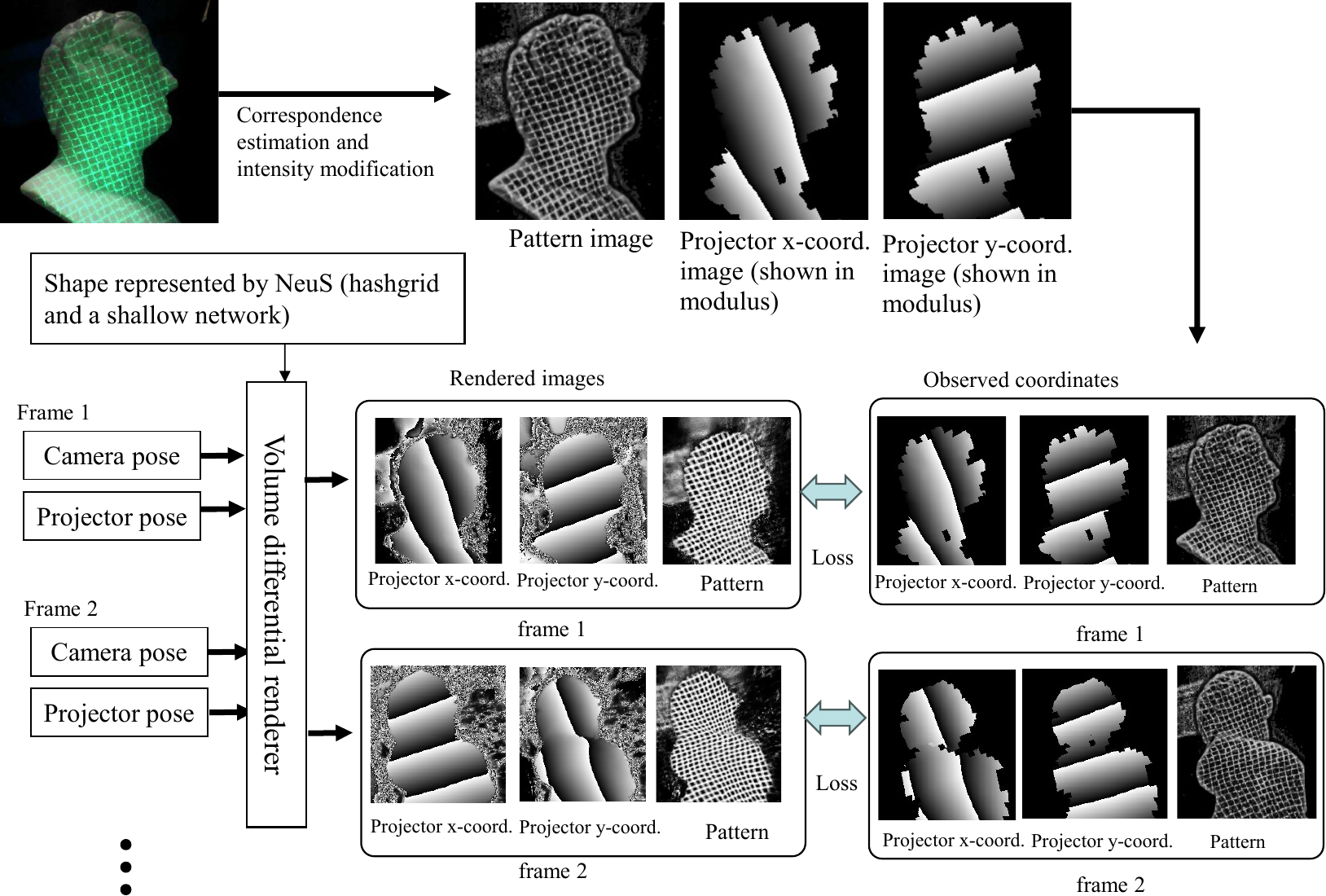}\\
        \vspace{-2mm}
(a)
     \end{minipage}
     \hspace{4mm}
%
     \begin{minipage}[b]{0.134\hsize}
        \centering
        \ifx
        \includegraphics[width=0.16\linewidth]{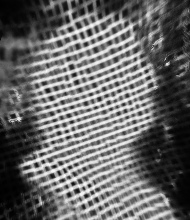}
        \includegraphics[width=0.16\linewidth]{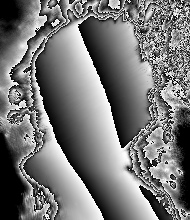}
        \includegraphics[width=0.16\linewidth]{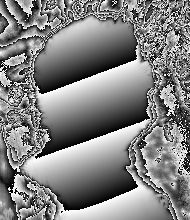}
        \includegraphics[width=0.16\linewidth]{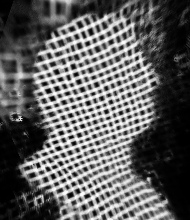}
        \includegraphics[width=0.16\linewidth]{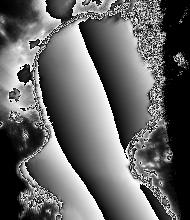}
        \includegraphics[width=0.16\linewidth]{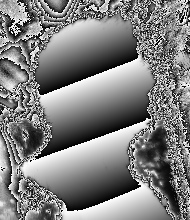}\\
        (a) \hspace{15mm} (b) \hspace{15mm}(c) 
        \hspace{15mm}(d) \hspace{15mm}(e) \hspace{15mm} (f)
        \caption{Effects of $s$ on pattern images (a, b),
        and $x$ and $y$ projector-coordinate images(b, c, e, f).
        For (a-c), $s=1/32$, and for (d-f) $s=1/72$. 
        Pattern images are rendered with blur for larger $s$ (\ie, (a-c)), 
        while coordinates images are less affected by $s$.}
        \fi
        \includegraphics[width=0.8\linewidth]{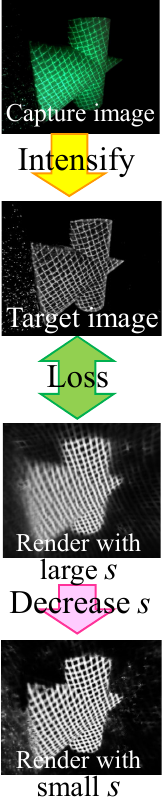}\\
(b)
     \end{minipage}
     \vspace{-2mm}
     \caption{(a) Multi-frame optimization using volumetric differential rendering. 
        (b) Effects of $s$ on pattern image. Leftmost and the second images are the captured 
         and the target image, and the third and rightmost images are rendered images with $s=1/32$, and $s=1/200$, respectively. 
        Pattern images are rendered with stronger blur for larger $s$, which results in that coordinates images are less affected by $s$.}
     \label{fig:multiframeoptimization}

\end{figure*}
     
     We optimize a neural surface representation 
     and camera/projector poses 
     by minimizing the discrepancy 
     between the rendered pattern images and the projector-coordinate images 
     towards respective target images. 
     For loss functions for the optimization, 
     we utilize L1 losses for the projector-coordinate images 
     and cosine losses for the pattern images. 
     This choice stems from the necessity to evaluate direct value differences 
     for projector-coordinate images 
     and to match the overall brightness distributions for pattern images. 
     The preference for using L1 loss over L2 loss is for robustness.
     
     Let the set of camera poses be 
     $M \equiv\{\mathbf{m}_1, \mathbf{m}_2, \cdots\}$,
     the set of projectors be $N\equiv\{\mathbf{n}_1, \mathbf{n}_2, \cdots\}$,
     where $\mathbf{m}_k$ represents camera pose parameters
     of frame $k$,
     $\mathbf{n}_k$ represents projector pose parameters
     of frame $k$,
     and the neural SDF be $f(x,y,z)$ as in equation \eqref{eqn:sdf}. 
     The cost function is
     \begin{align}
        L(f,M,N)   & \equiv w_{c} L_{c}(f,M,N) + w_{p} L_{p}(f,M,N) \nonumber            \\
        & + w_{e} L_{e}(f)\nonumber            \\
        L_{c}(f,M,N)  & \equiv \| \hat{\mathbf{C}} - \tilde{\mathbf{C}} \|_1  \nonumber \\
        L_{p}(f,M,N)  & \equiv \left ( 1 - \frac{
        \hat{\mathbf{P}}\cdot\tilde{\mathbf{P}}}{
        \sqrt{\hat{\mathbf{P}}\cdot\hat{\mathbf{P}}}
        \sqrt{\tilde{\mathbf{P}}\cdot\tilde{\mathbf{P}}}
        }  \right) \nonumber \\
        \hat{\mathbf{C}} & \equiv (\hat{\mathbf{C}}_1, \hat{\mathbf{C}}_2, \hat{\mathbf{C}}_3, \cdots) \nonumber \\
        \tilde{\mathbf{C}} & \equiv (\tilde{\mathbf{C}}_1, \tilde{\mathbf{C}}_2, \tilde{\mathbf{C}}_3, \cdots) \nonumber \\
        \hat{\mathbf{P}} & \equiv (\hat{P}_1, \hat{P}_2, \hat{P}_3, \cdots) \nonumber \\
        \tilde{\mathbf{P}} & \equiv (\tilde{P}_1, \tilde{P}_2, \tilde{P}_3, \cdots) 
     \end{align}
     where $\hat{\mathbf{C}}_i$ and $\hat{P}_i$ are rendered results of 
     pattern-coordinate images and pattern images for 
     the $i$-th sample pixel, 
     $\tilde{\mathbf{C}}_i$ and $\tilde{P}_i$ are target values for the same sample, 
     $\| \cdot \|_1$ is L1 loss,
     $\hat{\mathbf{C}}$ and 
     $\tilde{\mathbf{C}}$ are stacked vectors of 
     $\hat{\mathbf{C}}_i$ and 
     $\tilde{\mathbf{C}}_i$.
     $L$ is the total cost function, 
     $L_c$ is the L1 cost function 
     for coordinate images, 
     $L_p$ is the cost function 
     for pattern images,
     $L_e$ is Eikonal loss
     \cite{gropp2020implicit} to regularize SDF,
     $w_{c}$, $w_{p}$ and $w_{e}$are manually-defined weights
     for coordinate-image loss,
     pattern-image loss,
     and Eikonal loss.

     The loss function $L$ is minimized by differential rendering
     the Monte Carlo sampled pixels of projector-coordinate images 
     $\hat{\mathbf{C}}_1, \hat{\mathbf{C}}_2, \hat{\mathbf{C}}_3, \cdots$
     and
     pattern images 
     $\hat{P}_1, \hat{P}_2, \hat{P}_3, \cdots$, 
     calculating $L$, 
     back propagating $L$, 
     and updating $f$, $L$ and $M$ as shown in \figref{fig:multiframeoptimization}(a).

     When volumetric rendering from the neural-represented signed distance function, 
     the scale parameter $s$ of the sigmoid function determines 
     `thickness' of rendered surfaces. 
     A large value of $s$ results in integrating values 
     for long segment of 3D points along each ray, 
     leading to a `blurry' image in the rendering. 
     Therefore, $s$ is desirable to be small 
     in the last stages of the optimization. 
     However, if this value is too small from the initial stages, 
     only a small fraction of volume information for each ray is utilized, 
     leading to unstable optimization. 
     Thus, we start with a large value for $s$ and gradually decrease it during optimization.
     This process also works as a coarse-to-fine strategy, 
     where first blurry but global shape information is optimized, 
     and shaper but local shape information is optimized later. 
     
     The effect of value $s$ is more distinct 
     for pattern images,
     compared to projector-coordinate images as shown in \figref{fig:multiframeoptimization}(b). 
     For rendering projector-coordinate images, 
     projector coordinates are integrated, 
     which locally vary monotonously along each ray in the 3D space. 
     As a result, even if the integration range expands around the surface due to large $s$, 
     the output projector coordinates do not change much 
     (similar to applying moving average to monotonically increasing/decreasing functions). 
     In the case of the pattern image, 
     the values being integrated are the intensities of the pattern image. 
     In this case, the pattern becomes blurry for large $s$.
     
     By gradually decreasing $s$ in the optimization process
     pattern images are first blurry while projector-coordinate images are not much effected, 
     but they are rendered clearly in the later stages where $s$ is small. 
     Thus, 
     the L1 loss for projector-coordinate images is first utilized in the early stages,
     and cosine loss for pattern images is later emphasized. 
     \fi

We optimize the neural surface representation and the camera/projector poses by minimizing the discrepancy between rendered and target projector-coordinate and pattern images.

We use L1 loss for projector-coordinate images to evaluate direct value differences and cosine loss for pattern images to match brightness distributions. L1 is preferred over L2 for robustness.

Let the camera pose set be $M \equiv\{\mathbf{m}_1, \mathbf{m}_2, \cdots\}$ and the projector pose set be $N \equiv\{\mathbf{n}_1, \mathbf{n}_2, \cdots\}$, where $\mathbf{m}_k$ and $\mathbf{n}_k$ are the $k$-th frame’s camera and projector poses. Let $f(x,y,z)$ be the neural SDF defined in Eq.~\eqref{eqn:sdf}. The cost function is defined as:
    
          \begin{align}
        L(f,M,N)   & \equiv w_{c} L_{c}(f,M,N) + w_{p} L_{p}(f,M,N) \nonumber            \\
        & + w_{e} L_{e}(f)\nonumber            \\
        L_{c}(f,M,N)  & \equiv \| \hat{\mathbf{C}} - \tilde{\mathbf{C}} \|_1  \nonumber \\
        L_{p}(f,M,N)  & \equiv \left ( 1 - \frac{
        \hat{\mathbf{P}}\cdot\tilde{\mathbf{P}}}{
        \sqrt{\hat{\mathbf{P}}\cdot\hat{\mathbf{P}}}
        \sqrt{\tilde{\mathbf{P}}\cdot\tilde{\mathbf{P}}}
        }  \right) \nonumber \\
        \end{align}
where $\hat{\mathbf{C}}, \tilde{\mathbf{C}}$ are vectors of rendered and target projector coordinates, and $\hat{\mathbf{P}}, \tilde{\mathbf{P}}$ are those of the pattern images. $L_e$ is the Eikonal loss~\cite{gropp2020implicit} for SDF regularization. The weights $w_c$, $w_p$, and $w_e$ are set manually.

The loss function $L$ is minimized by performing differential rendering of Monte Carlo sampled pixels for projector-coordinate images 
$(\hat{\mathbf{C}}_1, \hat{\mathbf{C}}_2, \cdots)$ and pattern images 
$(\hat{P}_1, \hat{P}_2, \cdots)$, computing $L$, backpropagating it, and updating $f$, $M$, and $N$.

     \ifx
        \jptext{
           ここで，
           $S$は，
           点の集合ではなく，シーンの形状であることに注意する必要がある．
           passive stereoにおいては，
           $F$はカメラ$C$とは独立に存在する実在の3次元特徴点であり，
           $F$と$C$を独立に動かすことが物理的に想定できる．
           もし，$F$を動かさずにある一つのカメラ$c_1$を動かしたとすると，
           残りのカメラ$c_1,\cdots,c_N$では観測は変化しない．
           これに対し，
           active stereoにおいては，
           フレーム1でのプロジェクタ$p_1$を動かした時の観測値は，
           $p_1$からのレイ$r_1$を変化させ，
           $S$と$r_1$との交点位置を変化させ，
           $c_1$での観測値に影響を及ぼす．
     
           このことは，
           Loss関数のパラメータに対する勾配を利用して最適化を
           行うアルゴリズムにおいて，
           大きな影響がある．
           bundle adjustと同じように，
           ３次元特徴点を位置をパラメータ$F$として設定し，
           $F$をカメラ$C$及び，
           プロジェクタ$P$に投影した
           reprojection errorを
           loss関数とすることは，
           active stereoの過程の直接的なモデリングではない．
        }

        \jptext{
           active stereoのmulti frame optimizationには
           前節で説明した観測モデルの違いの他に，
           passive stereoとの大きな違いがある．
           passive stereoでは
           sparseな3次元特徴点$F$のある点$F_1$が，
           複数のフレームにまたがって観測される．
     
           これに対し，
           active stereoでは，
           複数のフレームで観測される，
           独立した特徴点は存在しない．
           前節で述べた，
           $P$からのレイと
           $S$の交点は，
           $P$が変化すると変化する．
           このため，
           これは，
           $C$とな独立に存在する$F$とは異なる．
           これは，active stereoが
           texturelessでも動作することの
           代償でもある．
           代わりに，
           シーン形状$S$が，
           フレーム間で共有されることが，
           異なるフレームでのカメラ，プロジェクタの位置を拘束する(\figref{fig:inter-frame-constraints}).
        }

        \jptext{
           前節では，active stereoとpassive stereoの観測モデルの
           違いについて述べた．
           本節では，
           inverse renderingによって
           active stereoのmulti-frame optimizationを行うための，
           観測誤差モデルを定義する．
     
           前節で述べた，プロジェクタによるパターンの投影は，
           CGではしばしば投影マッピングとして表現される．
           投影マッピングは，
           「ピクセルシェーダ」を利用すると簡単に実装できる．
           「ピクセルシェーダ」は，
           描画される各画素について，
           camera座標系での３次元点$V_w$を受け取る．
           $V_w$を，world-to-projectorの座標変換と
           プロジェクタの内部パラメータを用いて
           プロジェクタのパターン平面に投影し，
           投影された2次元座標$U$をテクスチャ座標としてマッピングに利用すると，
           投影マッピングが実現できる．
     
           この時，2次元座標$U$をそのまま画素値としてレンダリングを行うと，
           図\figref{fig:projector-xy-image-rendered}のような画像が得られる．
           この画像は，カメラ画像上の点から，
           対応するプロジェクタの２次元座標へのマッピングである．
           このマッピング情報を，
           カメラ位置$C$，
           プロジェクタ位置$P$，
           シーン形状$S$に対してレンダリングした画像を，
           $R(S,C,P)$と表記する．
     
           $R(S,C,P)$は，
           active stereoにおいて，
           観測パターンからプロジェクタ位置をピクセル毎に推定した
           場合に得られる情報と，本質的に同じものである．
           $R(S,C,P)$の情報は，
           active stereoにおける
           投影の過程を
           そのままモデリングしていることに注意してほしい．
           つまり，カメラ位置$C$，
           プロジェクタ位置$P$，
           シーン形状$S$がそれぞれ，あるいは同時に変化した時，
           $R(S,C,P)$の変化は，
           active stereoの観測モデルで
           期待される変化と一致する．
        }
     
        In this section, we define the observation error model
        for active stereo.
        As shown in \figref{fig:observation-model},
        the 2D coordinates of the camera
        and the projector are related via the scene surface $S$.
        For virtual, CG scene,
        the same information
        can be rendered as a projection mapping.
     \fi

     \jptext{
        $R(S,C,P)$を最適化に利用することで，
        active stereoシステムのMulti-frame optimizationが実現できる．
        手順は以下の通りである．
     
        Step1:フレームごとにパターンを投影した画像を解析し，
        cameraからprojectorへの密なマッピングを得る．
        このマッピングを画像化したものを，
        $M_i$とする．
        ただし，$i$はフレーム番号である．
     
        Step2:$M_i$から，フレーム内での自己キャリブレーションを
        行い，
        プロジェクタ・カメラ間の位置パラメータを求め，
        フレームごとの３次元形状$\sigma_i$を求める．
     
        Step3:$\sigma_i$ $(i=1,2,\cdots,N)$を荒く位置合わせし，
        カメラ位置，プロジェクタ位置の初期値$C_0$,$P_0$とする．
        これは，手動や，ICPアルゴリズムなどを利用する．
     
        Step4:初期値$C_0$で位置合わせされた
        $\sigma_i$ $(i=1,2,\cdots,N)$に近い，荒いメッシュ形状を
        作成し，初期形状$S_0$とする．
        これは，例えば$\sigma_i$ $(i=1,2,\cdots,N)$をメタボールで
        融合するなどの方法で実行できる．
     
        Step5:カメラ位置$C$(初期値$C_0$),
        プロジェクタ位置$P$(初期値$P_0$),
        シーン形状$S$(初期値$S_0$)から
        $R(S,C,P)$を計算する．
        その$i$フレーム目を
        $R_i(S,C,P)$とし，
        $L(S,C,p)=\sum_i \log( | R_i(S,C,P) - M_i |^2 + 1)$
        を損失関数として,
        $S$, $C$,$P$についての最適化を行う．
        $\tilde{S},\tilde{C},\tilde{P}=\argmin_{S,C,P} L(S,C,P)$
     }

     \jptext{
        我々は，
        inverse renderingによる
        全体最適化システムを，
        Pytorch3Dで実装した．
        Pytorch3Dのレンダリングシステムにおいて，
        各視点での描画時のパラメータとしてプロジェクタの位置を
        指定し，
        投影マッピングと同様にプロジェクタ側での２次元座標を計算し，
        それをピクセル値とする描画を行う．
        投影される座標の計算は，
        形状メッシュの頂点ごとに行い，
        メッシュのポリゴン内部での
        ピクセル値は
        頂点でのピクセル値の補間よって計算される．
     
        この処理を，各視点について行い，
        実測によって取得されたパターン座標のマップとの差から
        損失関数を計算した．
        損失関数をの最小化を，
        back propagationによって
        メッシュ頂点の位置，カメラ位置，プロジェクタ位置
        について行う．
     }

     \section{Experiments}\label{sec:experiments}

     
     

     \jptext{
        提案手法の検証のために，シミュレーションによる
        プロジェクタ・カメラ間での対応マップを作製した．
        \figref{fig:correspondence-map}は対応マップの例である．
        この例について，正解となるカメラ位置，プロジェクタ位置に
        乱数を加え，
        さらに，正解形状を強くsmoothingしたものを
        初期パラメータとして，
        提案手法を適用した．
     }
     

     \jptext{
        \figref{fig:exp-sim-bunny}に
        形状モデルの推定結果を示す．
        初期形状モデルには無い詳細形状が，
        提案手法によって復元されていることがわかる．
     
        また，既存手法との比較として，
        上記のように乱されたカメラ・プロジェクタ位置で奥行き情報を推定し，
        それをOpenCVのKinfuによって統合した．
        また，
        提案手法で修正した
        パラメータで推定された奥行き情報を利用して，
        Kinfuによって統合した．
        \figref{fig:exp-sim-bunny}に結果を示す．
        全体最適化により，形状間のスケールや
        consistencyが向上し，
        統合結果が改善していることがわかる．
     }


     \ifx
     \begin{figure}[t]
         \vspace{-3mm}
        \centering
        \includegraphics[height=22mm]{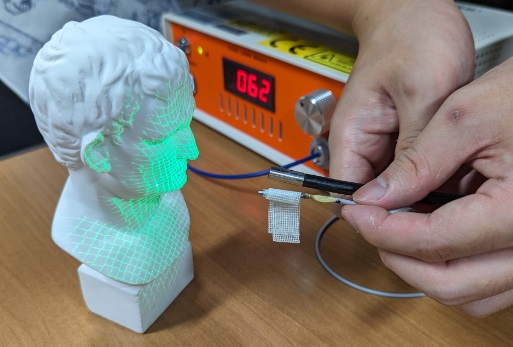} 
         \includegraphics[height=22mm]{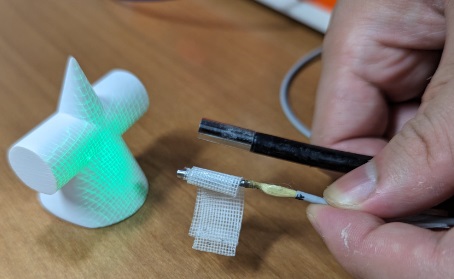}\\ 
         \vspace{-1mm}
          (a) \hspace{25mm} (b)\\
          \vspace{1mm}
        \includegraphics[height=22mm]{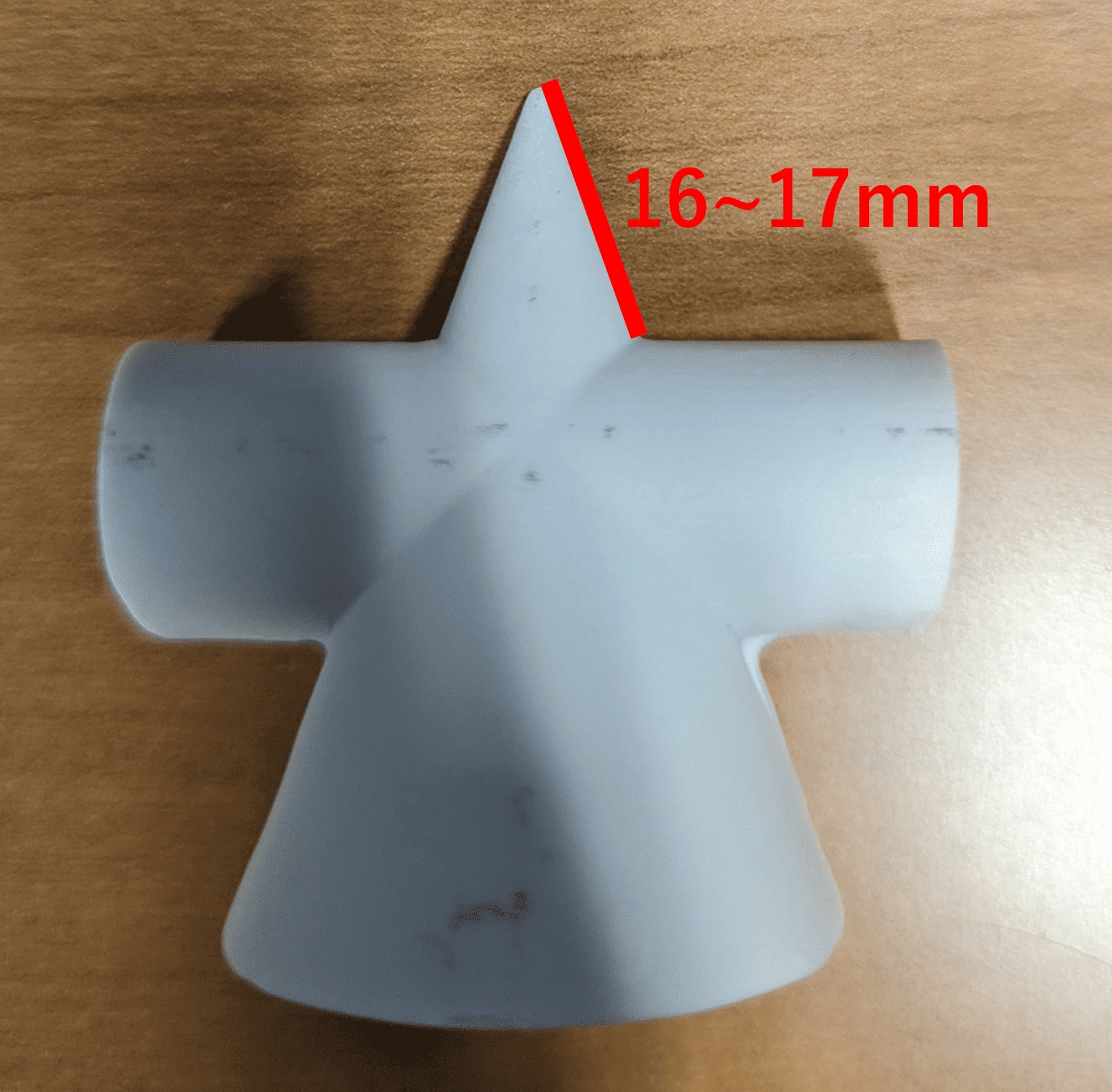} 
        \includegraphics[height=22mm]{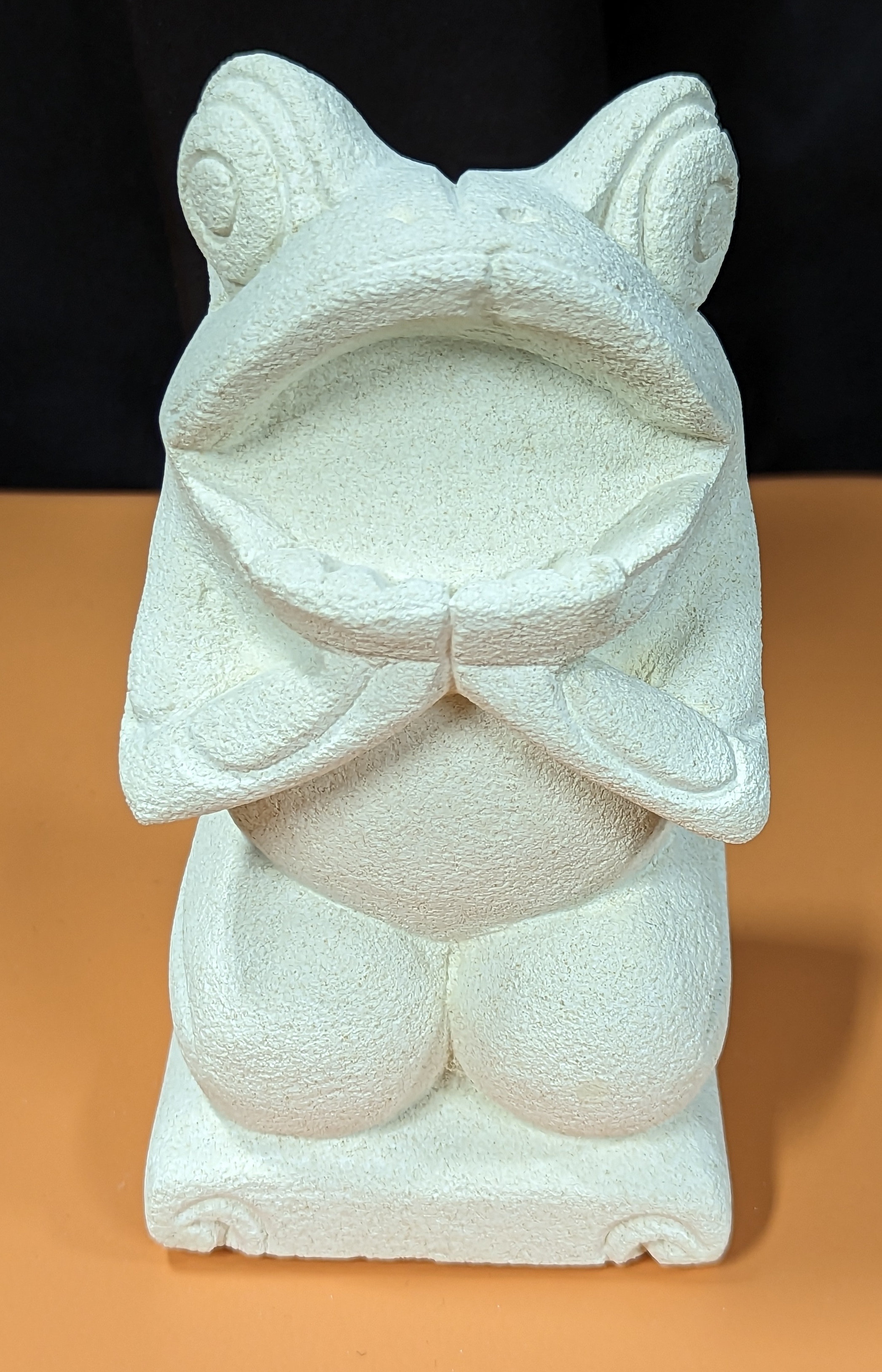} 
        \includegraphics[height=22mm]{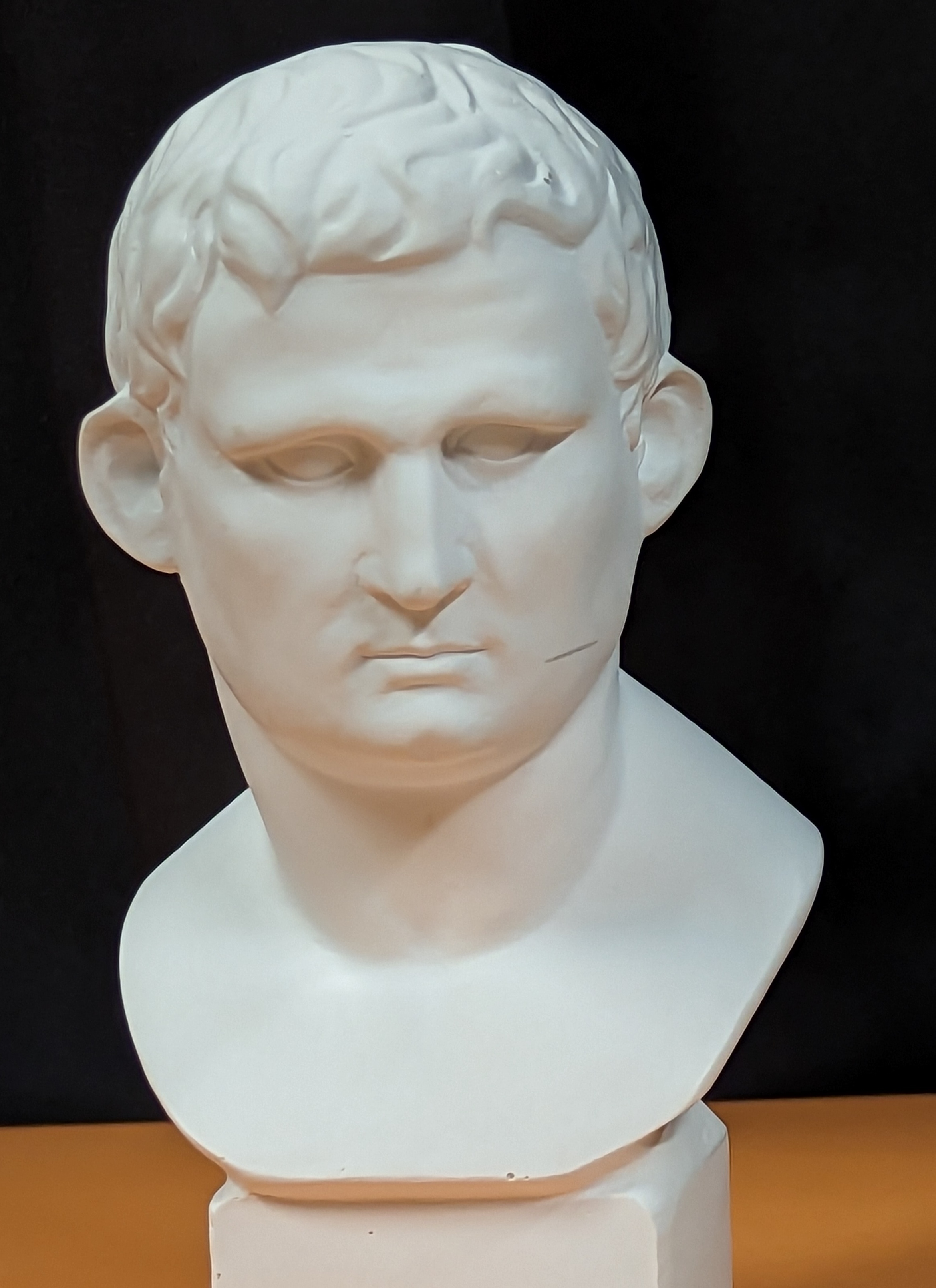}\\  
         \vspace{-1mm}
       \hspace{5mm}(c) \hspace{12mm} (d) \hspace{10mm} (e) \\ 
         \vspace{-2mm}
        \caption{Appearances of the experimental system and measured 3D objects.
        (a) and (b) Capturing with pattern projection, where a camera and a projector are freely moved during scan. (c) Appearance of `cone-and-cylinder' object.
     (d) Appearance of `frog' object.  (e) Appearance of `head' object.
        }
        \label{fig:capturesetup}
        \vspace*{-4mm}
     \end{figure}
     \fi 
        
     \ifx
     \begin{figure}[t]
         \vspace{-3mm}
        \centering
        \includegraphics[height=26mm]{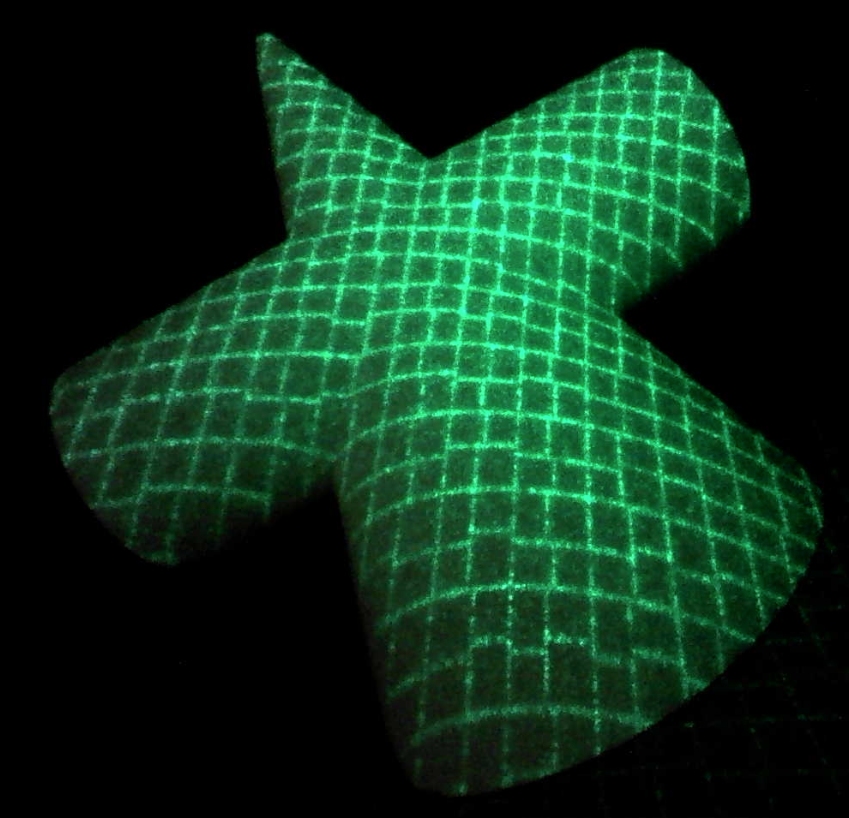} 
        \includegraphics[height=26mm]{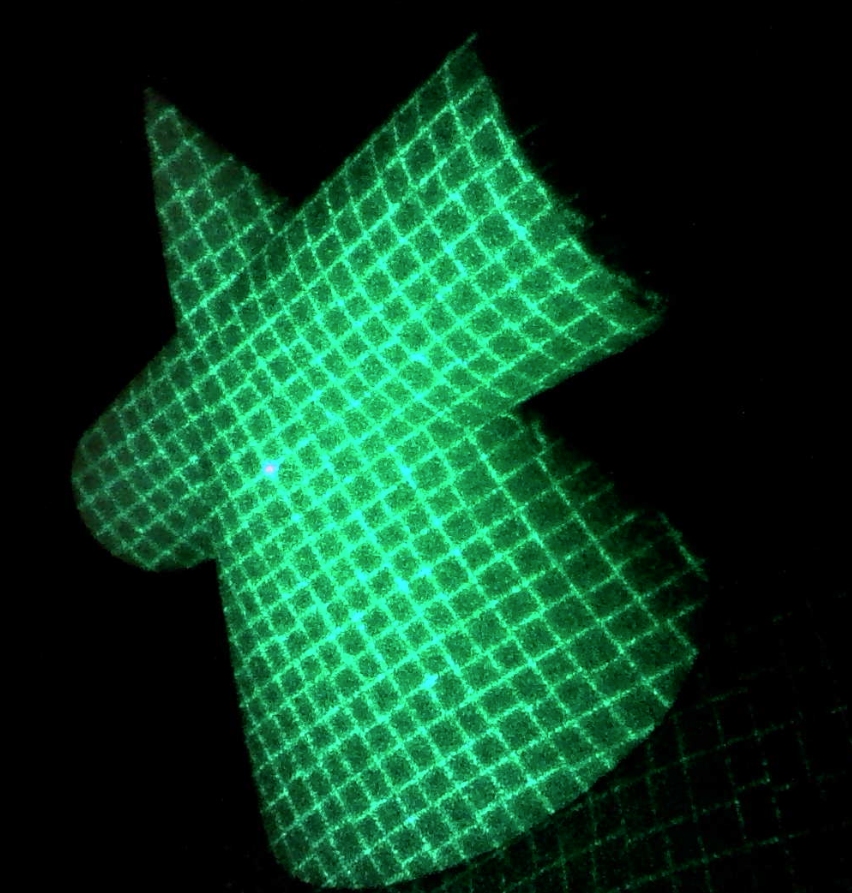} 
        \includegraphics[height=26mm]{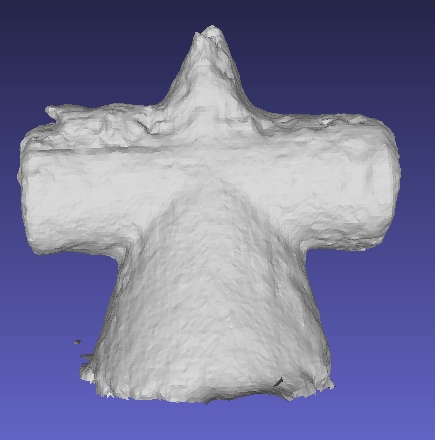}\\ 
        (a) \hspace{20mm}(b) \hspace{20mm}(c)  \\
        \includegraphics[height=29mm]{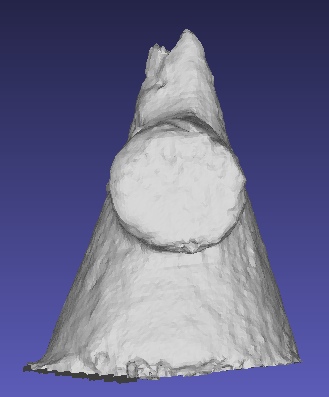}
        \includegraphics[height=29mm]{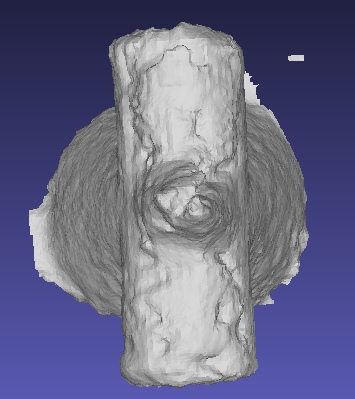} 
        \includegraphics[height=29mm]{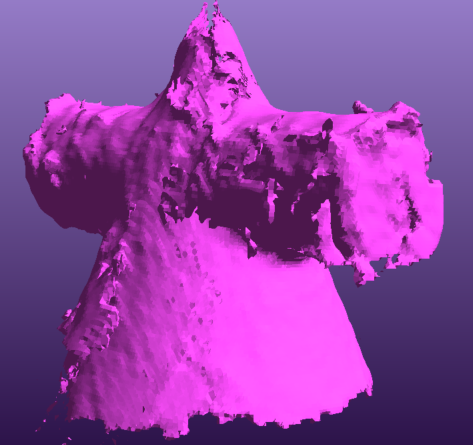} 
        (d) \hspace{20mm}(e) \hspace{20mm}(f) \\
        \vspace{-3mm}
        \caption{Shape reconstruction of `cone-and-cylinder' object.
        (a)(b) Captured images while moving both the camera and the projector. 
      (c-e) Shape reconstructed from 22 frames with the proposed method (5000 iterations).
     (f) Result of KinectFusion~\cite{newcombe2011kinectfusion}.}
        \label{fig:resultcone}
        \vspace*{-4mm}
     \end{figure}
     \fi 

     \subsection{Implementation details} \vspace*{-1mm}
     \ifx
     We implemented the proposed method 
     on the basis of Nerfstudio \cite{tancik2023nerfstudio}
     on a system with 16 GB of GPU memory.
     
     In the following experiments, 
     we used HashGrid of 
     tiny-cuda-nn library
     \cite{tiny-cuda-nn}
     for the neural representation of SDF. 
     The setup of the HashGrid was 5 levels of scales, 
     with the minimum scale of
     $2^4 \times 2^4 \times 2^4$, 
     the second minimum of
     $2^5 \times 2^5 \times 2^5$, $\cdots$,
     and the maximum resolution of
     $2^8 \times 2^8 \times 2^8$.
     For each 3D position, 
     the HashGrid of each scale level outputs 
     8-D feature;
     thus, the total dimension of the HashGrid output is $8 \times 5=40$
     
     The 40-D feature was processed by MLP with 2 hidden layers with 
     ReLU activation. The number of neurons for each hidden layer 
     was 128. 
     The output of the MLP was a 1-D SDF value for each 3D point. 
     
     
     For the optimization of the HashGrid, projector poses and camera poses, 
     we use multiple images 
     with the static pattern of \figref{fig:system}(b:top) projected onto surfaces. 
     The image resolution was $1200 \times 1200$. 
     For each target object, 
     we captured multiple images; 
     57 and 43 images for the 2 examples that are shown later. 
     
     The captured images were processed to obtain the x and y projector-coordinate for each image. 
     In the process, image regions where the pattern exists were predicted by U-Nets. 
     These regions of pattern existence were used as optimization masks. 
     Also, the pixel intensities were normalized for each local regions 
     to produce pattern images. 
     
     From the projector-coordinate images, 
     3D shapes were reconstructed for each frame using temporally-calibrated pose parameters.
     The shapes were automatically aligned using ICP algorithm~\cite{besl1992method}.
     The aligned shape poses were used for initial pose parameters for the optimization. 
     
     For the image set of the 
     projector-coordinate images 
     and pattern images, 
     we sampled 2024 pixels (Monte Carlo sampling)
     for each iteration, 
     and volume-rendered the projector coordinates and pattern intensities for those pixels. 
     We sampled 64 voxels for each ray.
     We calculate L1 loss for the projector coordinates (dimension being $2024 \times 2$) and 
     cosine loss for the pattern intensities (dimension being 2024). 
     Based on sum of these losses, 
     HashGrid, projector poses and camera poses were updated. 
     These processes form a single iteration of optimization.

     For scale parameter $s$, $\frac{1}{s}$ was initially set to 
     $32.0$, 
     and $\frac{1}{s}$ was increased by 0.02 for each iteration
     (\ie, decreasing $s$).
     After $\frac{1}{s}$ reaches 200, 
     $\frac{1}{s}$ was kept constant,
     since, if $s$ becomes too small, 
     the optimization becomes unstable because of shortage of 
     sampling around the surface.
     
     We used learning rate decay. 
     For the SDF field, 
     max learning learning rate was 
     $5.0 \times 10^{-4}$
     and gradually decreased the rate to 
     $5.0 \times 10^{-12}$ at 8000 iteration.
     For the camera and the projector pose parameters, 
     learning max learning rate was 
     $1.0 \times 10^{-3}$
     and gradually decreased the rate to 
     $5.0 \times 10^{-12}$ at 8000 iteration.
     These criteria were decided empirically. 
     
     
     In the experiments, we repeated the optimization process up to 10000 iterations.
     We set $w_c$ and $w_p$ such that 
     $w_c~L_c$ and $w_p~L_p$ becomes similar values
     in the final stages of optimization. 
     In our case, 
     $w_c=1000$, $w_p=0.05$.
     $w_e$ was set such that 
     $W_e~L_e$ becomes about $\frac{1}{10}$ of 
     $w_c~L_c$.
     In our case, 
     $w_e=0.01$.
     
     A typical execution time was about
     1 min for 1000 iterations of optimization step.
     \fi     

We implemented the proposed method based on Nerfstudio~\cite{tancik2023nerfstudio} and ran it on a GPU with 16\,GB memory.

We employed the HashGrid of the tiny-cuda-nn library~\cite{tiny-cuda-nn} with 5 resolution levels ranging from $2^4$ to $2^8$ per axis. Each level outputs an 8-dimensional feature, resulting in a 40-D vector. This was input to an MLP with two hidden layers (128 neurons, ReLU) to predict 1D SDF values.

We used images of resolution $1200 \times 1200$, captured under a static projected pattern (\figref{fig:system}(b:top)). For two objects, we captured 57 and 43 images. Projector-coordinate images were estimated using U-Nets, and pattern regions were used as optimization masks. Local intensity normalization was applied to generate pattern images.

Initial camera and projector poses were computed by reconstructing 3D shapes from projector-coordinate images, followed by alignment using ICP~\cite{besl1992method}. These were used as initialization for optimization.

In each iteration, 2024 pixels were sampled, and 64 voxels per ray were volume-rendered. L1 loss (for projector coordinates) and cosine loss (for pattern images) were computed and used to update the SDF, camera, and projector parameters.



We set $w_c = 1000$, $w_p = 0.05$, and $w_e = 0.01$, so that $w_c L_c$ and $w_p L_p$ had similar magnitudes, and $w_e L_e$ was about one-tenth of $w_c L_c$ in the final stage.
A typical execution time was about 1 minute per 1000 iterations.

     \begin{figure}[t]
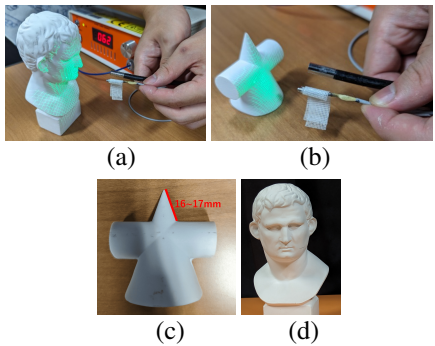

         \vspace{-3mm}
        \centering
        \includegraphics[height=18mm]{figs-icra25/cap_statue.jpg} 
         \includegraphics[height=18mm]{figs-icra25/cap_cone.jpg}\\ 
         \vspace{-1mm}
          (a) \hspace{20mm} (b)\\
          \vspace{1mm}
        \includegraphics[height=18mm]{figs-icpr24/conecylinder.jpg} 
        \includegraphics[height=18mm]{figs-icpr24/statue-appearance01.jpg}\\  
         \vspace{-1mm}
       \hspace{5mm}(c) \hspace{12mm} (d) \\ 
         \vspace{-2mm}
        \caption{Appearances of the experimental system and measured 3D objects.
        (a) and (b) Capturing with pattern projection, where a camera and a projector are freely moved during scan. (c) Appearance of `cone-and-cylinder' object.
     (d) Appearance of `head' object.
        }
        \label{fig:capturesetup}
        \vspace*{-4mm}
     \end{figure}

     \subsection{Comparison with other techniques} \vspace*{-1mm}

    \ifx
     First, we captured 3D shapes of real objects by moving either an object, a projector or a camera 
     as shown in \figref{fig:capturesetup}(a)(b), and reconstructed the shapes using 
     by our technique.
     The target object is statues of 
     'cone-and-cylinder' as shown in \figref{fig:capturesetup}(c).
     We captured 22 images while moving both the camera positions 
     and the projector positions, as shown in \figref{fig:resultcone}(a) and (b).
     The reconstruction results 
     are shown in \figref{fig:resultcone}(c-e). 
     We also capture the same object using KinectFusion~\cite{newcombe2011kinectfusion} as for the ground truth. 
     As the figure shows, we could confirm that the proposed method performed better than KinectFusion~\cite{newcombe2011kinectfusion} (shown in \figref{fig:resultcone}(f)).
     We also calculated RMSE of cone-and-cylinder shape and it was 0.97mm for KinectFusion, whereas 0.619mm for our method.
     When we carefully see the results,
     shapes from the multiple images were 
     merged naturally, 
     although the sharp shape around the 
     conical vertex was smoothed out. 
     It is because decoding projected pattern 
     for sharp-edged regions is difficult, 
     and also because SDF representation for such
     sharp-edged regions are difficult.  
     \fi 

We evaluated our method by reconstructing 3D shapes of real objects while moving the camera, the projector, or the object, as shown in \figref{fig:capturesetup}(a)(b). The target was a 'cone-and-cylinder' statue (\figref{fig:capturesetup}(c)). We captured 22 images under varying camera and projector positions.

Reconstruction results using our method are shown in \figref{fig:resultcone}(a-c). For comparison, we also scanned the same object using KinectFusion~\cite{newcombe2011kinectfusion}, treated as ground truth (\figref{fig:resultcone}(d)). As shown, our method provided more accurate shape recovery, with an RMSE of 0.619\,mm compared to 0.97\,mm by KinectFusion.

The integration of multiple views was performed smoothly. However, the conical vertex was slightly smoothed due to difficulties in decoding projected patterns and representing sharp edges with the SDF. Despite this, our method effectively merged observations and preserved global shape structure better than the baseline.

     \begin{figure}[t]
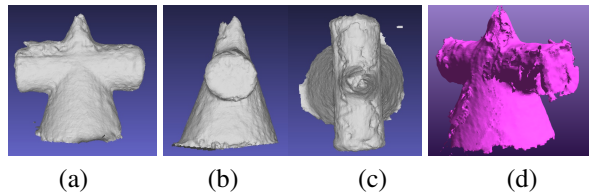

         \vspace{-3mm}
        \centering
        \includegraphics[height=20mm]{figs-icra25/cone-reconst01.jpg} 
        \includegraphics[height=20mm]{figs-icra25/cone-reconst02.jpg}
        \includegraphics[height=20mm]{figs-icra25/cone-reconst03.jpg} 
        \includegraphics[height=20mm]{figs-wacv25/kinfu-cone0.png}\\ 
        (a) \hspace{15mm}(b) \hspace{15mm}(c) \hspace{15mm}(d) \\
        \vspace{-3mm}
        \caption{Shape reconstruction of `cone-and-cylinder' object.
      (a-c) Shape reconstructed from 22 frames with the proposed method (5000 iterations).
     (d) Result of KinectFusion~\cite{newcombe2011kinectfusion}.}
        \label{fig:resultcone}
        \vspace*{-4mm}
     \end{figure}
     
     
     
     \ifx
     \begin{figure}[t]
        \centering
        \includegraphics[height=20mm]{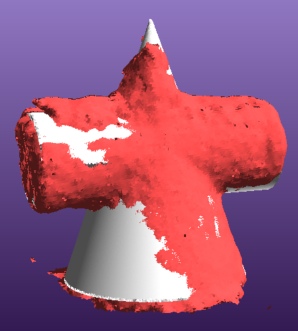} %
        \includegraphics[height=20mm]{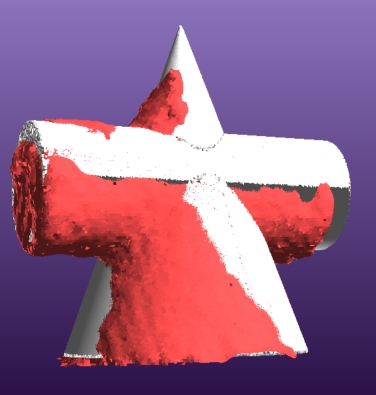}
        \includegraphics[height=20mm]{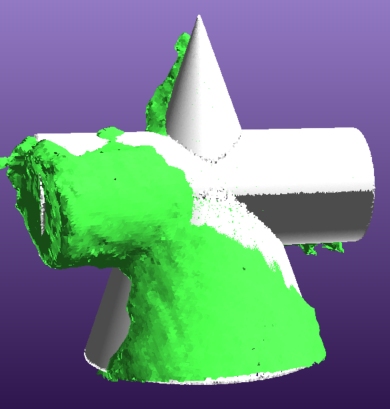}
        \includegraphics[height=20mm]{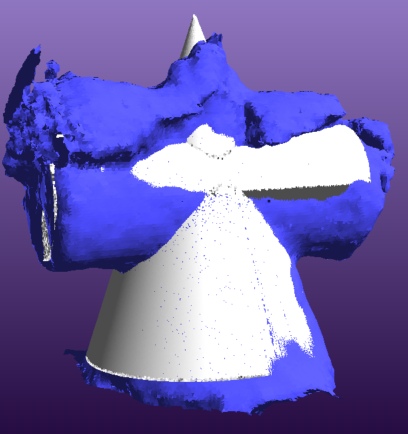}\\
        (a) \hspace{15mm}(b)\hspace{15mm} 
        (c) \hspace{15mm}(d) 
        \vspace{-3mm}
        \caption{Aberration studies.
        Cases are
        (a) Proposed method,
        (b) the result without using pattern-image loss,
        (c) the result without optimizing projector/camera poses,
        and (d) the result without decreasing scale 
        parameter $s$.
        The reconstructed shapes were aligned with Ground-truth shape (the white shapes in the figures) with an ICP method. 
        }
        \label{fig:aberration}
     %
     \vspace*{2mm}
     \makeatletter
     \def\@captype{table}
     \makeatother
     %
        \caption{ICP RMSEs for results in 
        \figref{fig:aberration}.
        }
         \label{tab:aberration}
     \begin{tabular}{|r|r|r|r|r|} 
         \hline
       & (a) & (b)  w/o & (c)  w/o &(d)  w/o \\ 
       & all &pat loss & pose opt. & scale\\  \hline
      ICP errors(mm) & 5.94& 9.11 & 9.33 & 9.80 \\
       \hline
      \end{tabular}
     \vspace{-4mm}
     \end{figure}
     \fi 

     
     
     \subsection{Ablation study} \vspace*{-1mm}
     \ifx
     Next, we conducted ablation studies 
     to show effectiveness of the 
     components of the proposed method.
     The tested components were
     pattern-image loss,
     projector/camera pose optimization, 
     and decreasing scale parameter $s$ while
     optimization.
     We reconstructed the example without using 
     those features.
     The results were
     aligned with the GT shape with ICP. 
     \figref{fig:aberration} and 
     \tabref{tab:aberration} show the results.
     As the results show, 
     without using pattern-image loss,
     the result shape became much worse. 
     It is because, for many of the observed images, 
     large area of the projector coordinates were
     missing because of decoding errors. 
     Without projector/camera pose estimation, 
     the result shape became much worse 
     because of the errors of the 
     projector and camera poses that were previously
     aligned by ICP. 
     Without decreasing scaling parameter $s$, 
     the shape became over-smoothed.
     \fi 

     Next, we conducted ablation studies 
     to show effectiveness of the 
     components of the proposed method.
     The tested components were
     pattern-image loss,
     projector/camera pose optimization, 
     and decreasing scale parameter $s$ while
     optimization.
     We reconstructed the example without using 
     those features.
     The results were
     aligned with the GT shape with ICP. 
     \tabref{tab:aberration} show the results.
     As the results show, 
     without using pattern-image loss,
     the result shape became much worse. 
     It is because, for many of the observed images, 
     large area of the projector coordinates were
     missing because of decoding errors. 
     Without projector/camera pose estimation, 
     the result shape became much worse 
     because of the errors of the 
     projector and camera poses that were previously
     aligned by ICP. 
     Without decreasing scaling parameter $s$, 
     the shape became over-smoothed.

         \begin{figure}[t]
     %
     \makeatletter
     \def\@captype{table}
     \makeatother
     %
        \caption{ICP RMSEs for results.
        }
    \vspace{-2mm}
         \label{tab:aberration}
     \begin{tabular}{|r|r|r|r|r|} 
         \hline
       & (a) & (b)  w/o & (c)  w/o &(d)  w/o \\ 
       & all &pat loss & pose opt. & scale\\  \hline
      ICP errors(mm) & 5.94& 9.11 & 9.33 & 9.80 \\
       \hline
      \end{tabular}
     \end{figure}

     \ifx
     \begin{figure}[t]
        \begin{minipage}{0.23\columnwidth}
            \centering
                \includegraphics[width=0.8\columnwidth]{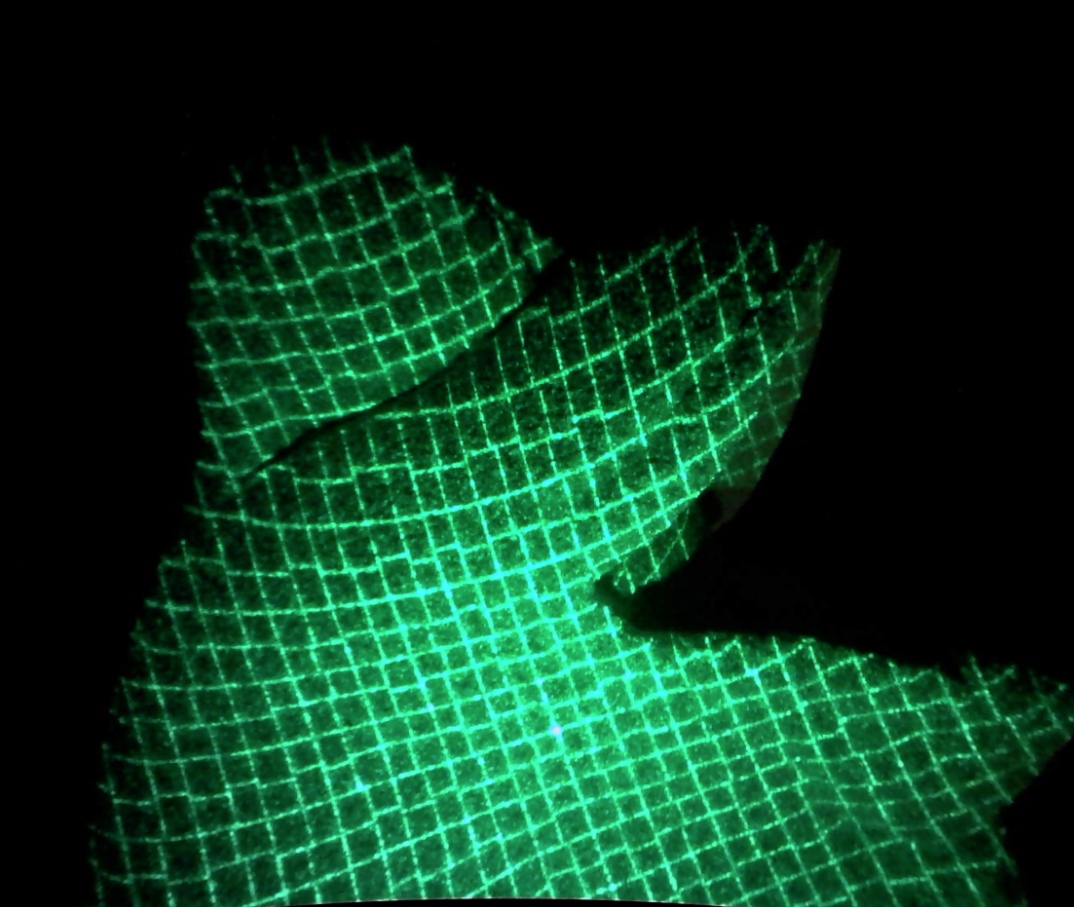} \\
                \includegraphics[width=0.8\columnwidth]{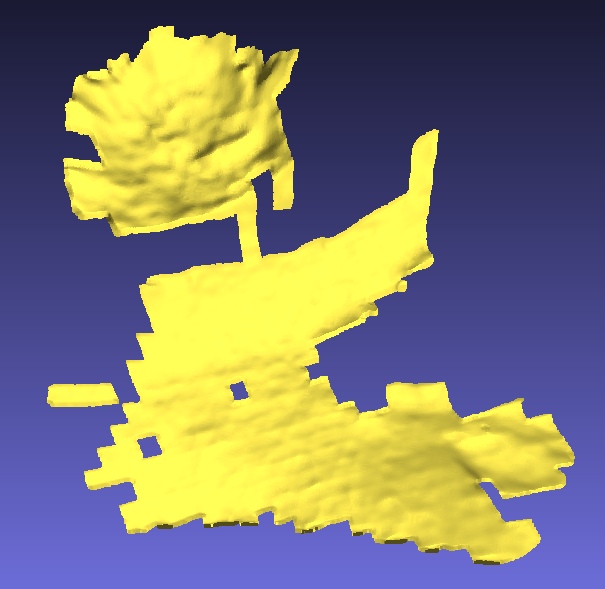}\\
                (a)
        \end{minipage} \hfill
        \begin{minipage}{0.76\columnwidth}
            \centering
            \includegraphics[height=29mm]{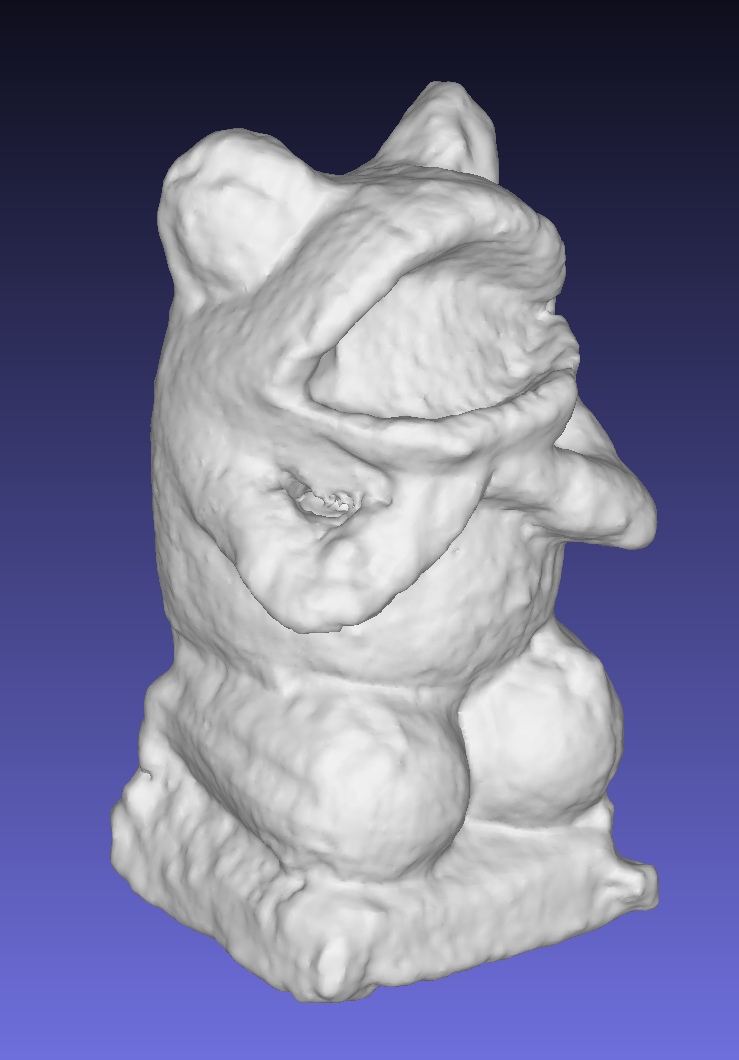} 
            \includegraphics[height=29mm]{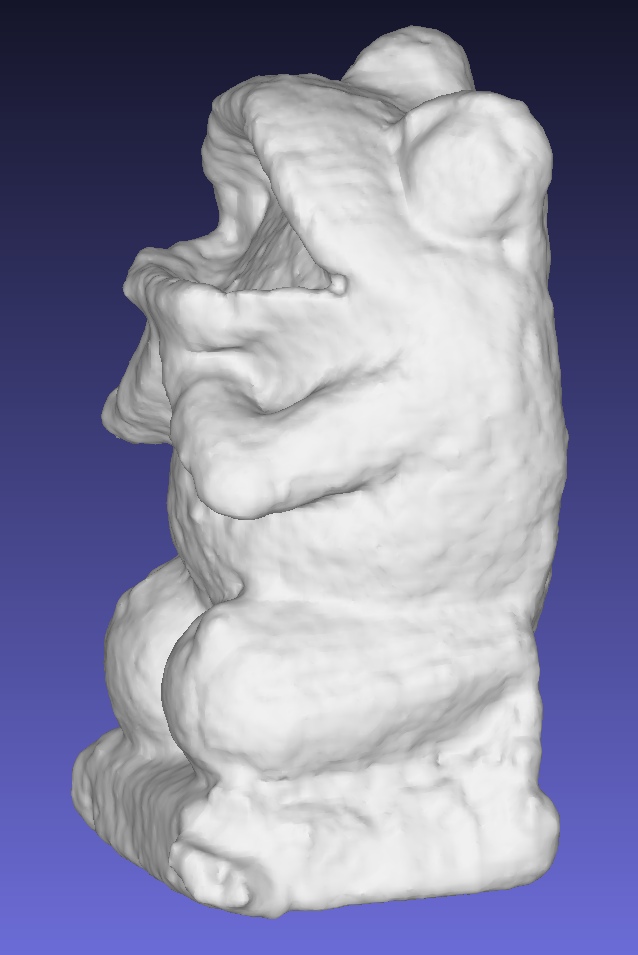} 
            \includegraphics[height=29mm]{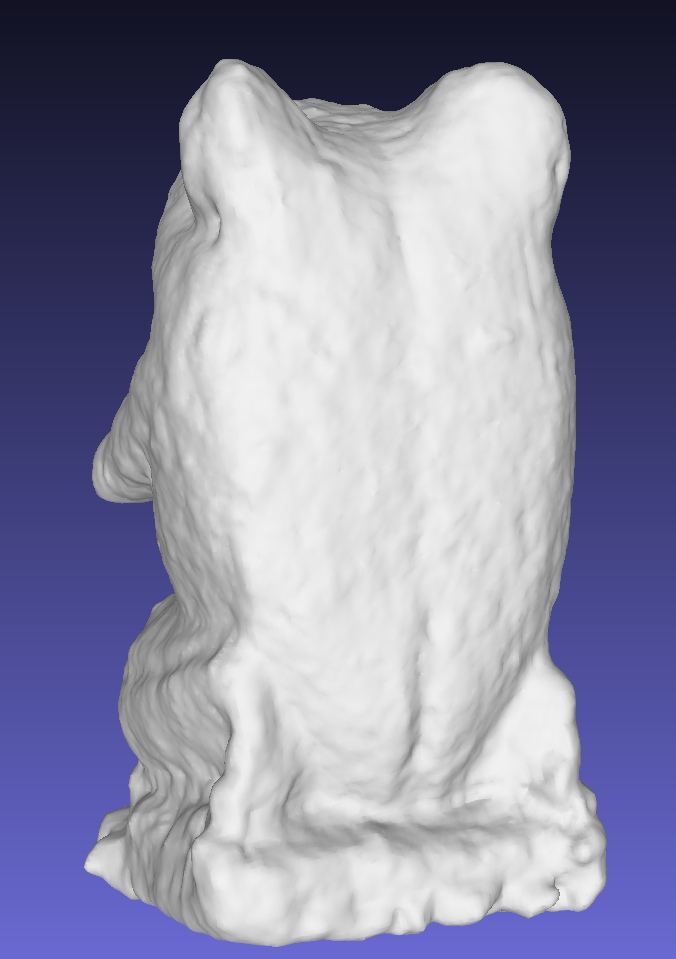} \\
            (b) \hspace{18mm}(c) \hspace{18mm}(d) \\
        \end{minipage}        
        \vspace{-3mm}
        \caption{Multi-frame optimization of `frog' object (57 frames).
        (a)Example of 1-frame reconstruction(captured image and the reconstructed shape).
        (b-d) Extracted shape
        after 10000 iterations.}\label{fig:resultfrog}
     \end{figure}
     \fi 
        
     \begin{figure}[t]
         \vspace{-2mm}
        \begin{minipage}{0.20\columnwidth}
            \centering
                \includegraphics[width=12mm]{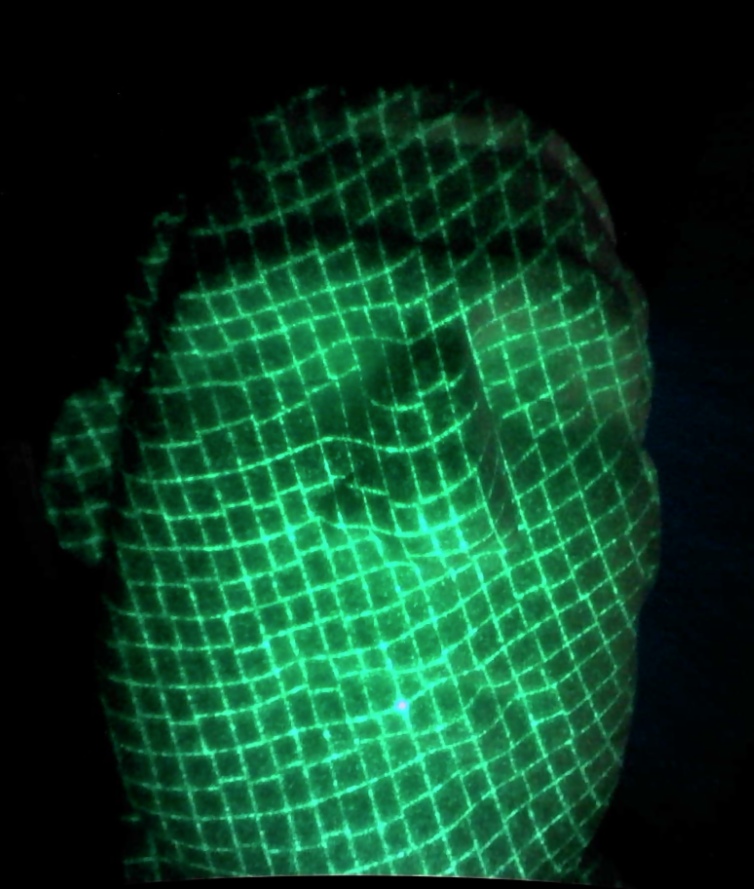} \\
                \includegraphics[width=12mm]{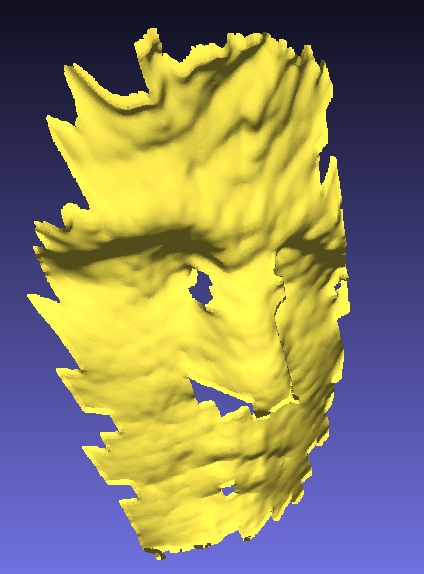} \\ %
                (a)
        \end{minipage} \hfill
        \begin{minipage}{0.79\columnwidth}
            \centering
            \includegraphics[height=31mm]{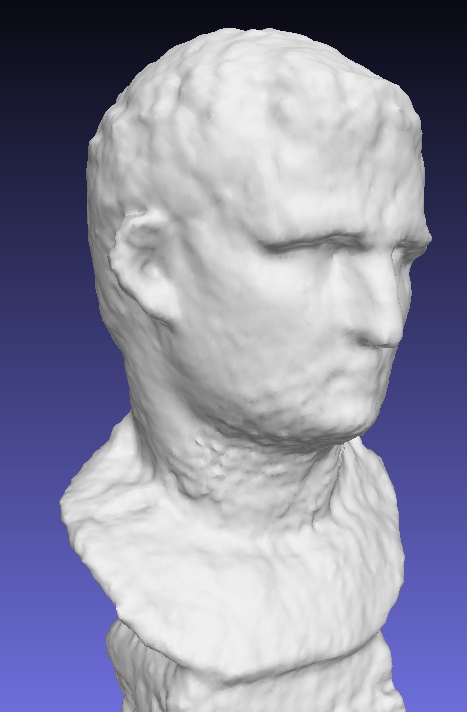} 
            \includegraphics[height=31mm]{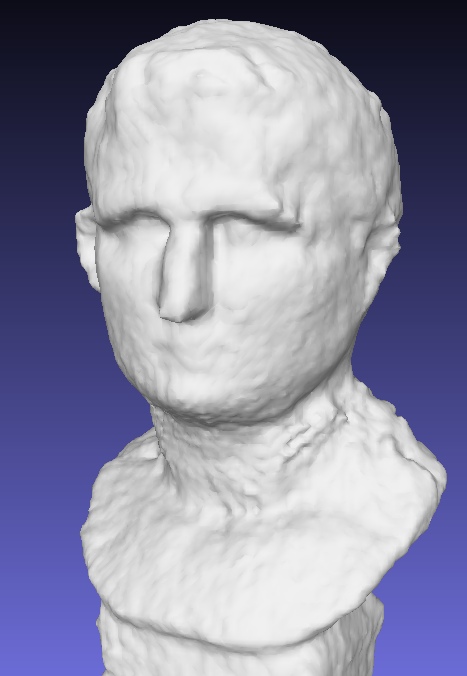} 
            \includegraphics[height=31mm]{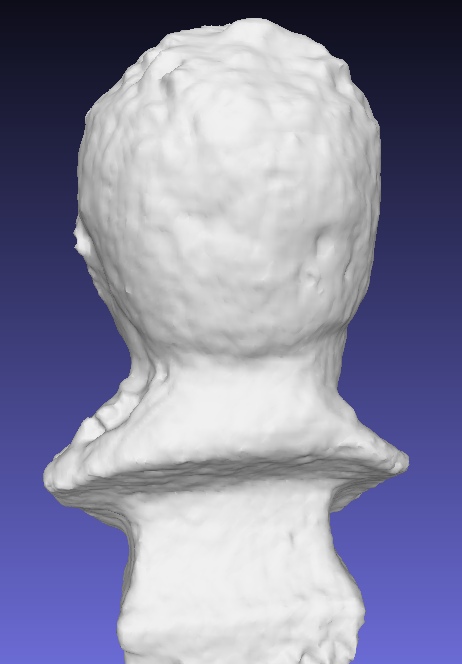}\\ 
             (b) \hspace{18mm}(c) \hspace{18mm}(d) \\
        \end{minipage} \hfill

        \vspace{-3mm}
        \caption{Results of multi-frame optimization. 
        (a)Example of 1-frame reconstruction.
        (b-c) A shape of `head' object reconstructed from 43 frames
        after 10000 iterations.}
        \label{fig:resulthead}
        \vspace*{-4mm}
     \end{figure}
          
     \subsection{Demonstration} \vspace*{-1mm}
     \ifx
     Finally, we reconstructed 
     The target objects are statues of `frog' and `head' as shown in \figref{fig:capturesetup}(d),(e).
     We captured 57 and 43 images for 
     frog and head, respectively.
     Examples of captured images and 
     1-frame reconstruction results are shown in
     \figref{fig:resultfrog}(a) and \figref{fig:resulthead}(a). 
     %
     The final reconstruction results 
     are shown in \figref{fig:resultfrog}(b-d) and \figref{fig:resulthead}(b-d).
     We can confirm that the entire shape of objects are successfully reconstructed by our method.
     \fi 

Finally, we demonstrated our method on a statue of a `head' (\figref{fig:capturesetup}(d)). We captured 43 images.
An example of captured images and a single-frame reconstruction is shown in \figref{fig:resulthead}(a). The final multi-frame result is shown in \figref{fig:resulthead}(b-d). This confirms that our method successfully reconstructed the complete shape of the object.

     
     \section{Conclusion}
     

     \ifx
     In the paper, a novel multi-frame 3D reconstruction method 
     based on neural shape representation 
     specialized for structured-light scanning was proposed. 
     In the method, 
     structured-light information such as camera-to-projector correspondences 
     or projected pattern intensities are 
     rendered by differential volume renderer. 
     The optimization can be done not only with respect to the scene shape information, 
     but also to the projector and camera poses; 
     thus, initial alignment errors between frames can be corrected. 
     \fi 

We proposed a novel multi-frame 3D reconstruction method using neural shape representation specialized for structured-light scanning. Structured-light cues, including camera-to-projector correspondences and projected pattern intensities, are rendered via differential volume rendering. Our framework jointly optimizes shape and device poses, enabling correction of initial alignment errors across frames.
     The proposed method was confirmed to work properly
     with the scanned data for datasets capturing real objects. 


\ifx 
\section*{ACKNOWLEDGMENT}
This work was supported by JST Startup JPMJSF23DR and JSPS/KAKENHI JP20H00611, JP23H03439 and NEDO(JPNP20006) in Japan.
\fi




{
        \small
  

     }

\end{document}